%% file: main.tex
\documentclass[10pt,twocolumn,letterpaper]{article}

\usepackage{cvpr}              

\input{preamble}

\definecolor{cvprblue}{rgb}{0.21,0.49,0.74}
\usepackage[pagebackref,breaklinks,colorlinks,citecolor=cvprblue]{hyperref}

\title{{RaCo}: Ranking and Covariance for Practical Learned Keypoints}

\author{
    Abhiram Shenoi$^1$ \hspace{1.5em} Philipp Lindenberger$^1$ \hspace{1.5em} Paul-Edouard Sarlin$^2$ \hspace{1.5em} Marc Pollefeys$^{1,3}$ \\
    \vspace{0.5em}
    $^1$ETH Zurich \hspace{2em} $^2$Google \hspace{2em} $^3$Microsoft Mixed Reality \& AI Lab
}

\begin{document}
\maketitle
\input{sec/0_abstract}

\input{sec/4_related_work}
\input{sec/5_method}
\input{sec/6_experiments_results}
\input{sec/7_conclusion}

\input{sec/X_suppl}

\clearpage

{
    \small
    \bibliographystyle{ieeenat_fullname}
    \bibliography{main}
}

\end{document}

%% file: preamble.tex
\usepackage[dvipsnames]{xcolor}

\usepackage[table]{xcolor}
\usepackage{multirow}
\usepackage{rotating}
\usepackage{makecell}
\usepackage{siunitx}
\usepackage{bm}
\usepackage{makecell}  
\usepackage{textcomp,gensymb}  
\usepackage{microtype}

\definecolor{tabfirst}{rgb}{1, 0.7, 0.7} 
\definecolor{tabsecond}{rgb}{1, 0.85, 0.7} 
\definecolor{tabthird}{rgb}{1, 1, 0.7} 

\newcommand{\0}{\phantom{0}}

\renewcommand{\paragraph}[1]{\vskip3pt \noindent\textbf{#1}}

\newcommand{\ours}{RaCo}

%% file: sec/0_abstract.tex
\begin{abstract}

This paper introduces \textbf{\ours}, a lightweight neural network designed to learn robust and versatile keypoints suitable for a variety of 3D computer vision tasks. The model integrates three key components: the repeatable keypoint detector, a differentiable ranker to maximize matches with a limited number of keypoints, and a covariance estimator to quantify spatial uncertainty in metric scale.
Trained on perspective image crops only, RaCo operates without the need for covisible image pairs. It achieves strong rotational robustness through extensive data augmentation, even without the use of computationally expensive equivariant network architectures. The method is evaluated on several challenging datasets, where it demonstrates state-of-the-art performance in keypoint repeatability and two-view matching, particularly under large in-plane rotations.
Ultimately, RaCo provides an effective and simple strategy to independently estimate keypoint ranking and metric covariance without additional labels, detecting interpretable and repeatable interest points.
The code is available at: \url{https://github.com/cvg/RaCo}

\end{abstract}

%% file: sec/4_related_work.tex
 \section{Introduction}

Sparse interest points are a key building block for large-scale 3D computer vision systems, enabling applications like 3D reconstruction~\cite{schoenberger2016sfm,agarwal2011building} and visual localization~\cite{sarlin2019coarse,sattler2018benchmarking}. 
A keypoint is a 2D point on a distinctive image region that can be reliably detected 
from various viewpoints and under different appearance changes,
enabling multi-view association. 
A set of keypoints form a sparse representation of an image, which reduces repetitive computations in downstream algorithms and thus increases their scalability.
Initially estimated via heuristics on low-level image statistics~\cite{harris1988combined,lowe2004distinctive,rublee2011orb}, modern data-driven approaches have proven to be significantly more robust~\cite{superpoint,dusmanu2019d2,tyszkiewicz2020disk,revaud2019r2d2}.
These detectors are trained using either synthetic data with ground-truth labels~\cite{superpoint} or self-supervision objectives without ground-truth labels~\cite{tyszkiewicz2020disk,zhao2023aliked}.
One then typically finds correspondences between keypoints across images using local descriptors.

\input{figures/teaser/fig}
 
The community has recently evaluated the performance of keypoint detection and description jointly on downstream tasks that are based on correspondences, such as relative pose estimation~\cite{nister2004efficient} and Structure-from-Motion (SfM)~\cite{imwchallenge2021}. 
This however conflates the performance of both steps.
While deep learning has greatly enhanced the robustness of descriptors, 
keypoint detection has not improved at the same rate:
classical algorithms, like SIFT~\cite{lowe2004distinctive},
remain highly competitive, especially in terms of orientation invariance and localization accuracy~\cite{sarlin2023pixsfm,santellani2023strek}.
One reason for this disparity is the difficulty in obtaining high-quality supervision for keypoints, which are poorly defined, compared to the relative ease in obtaining ground-truth correspondences~\cite{li2018megadepth,dai2017scannet}.
Furthermore, recent improvements in dense feature matching~\cite{edstedt2023roma,edstedt2023dkm,truong2021pdcnet} have made feature descriptors less important.
Keypoints, however, remain critical to create discrete multi-view tracks for SfM and scale to large scenes.
This thus results in new requirements for keypoint detection: 
keypoints should be highly precise and repeatable, but also agnostic to the matching that they are paired with.
Consequently, we evaluate keypoints in isolation and introduce a keypoint detector that is tailored for the challenges of the deep learning era, illustrated in~\cref{fig:teaser},
by exhibiting the following key qualities:

 \paragraph{Rotation Robustness:}
Simple in-plane rotations of images can cause both detections and correspondences to break catastrophically~\cite{santellani2023strek,bokman2024steerers}.
This has long been overlooked because such cases were not covered in existing datasets, but has recently gained increased attention~\cite{santellani2023strek,lee2022self,bokman2024steerers}. We show that careful rotation augmentations during training suffices to obtain competitive rotational detection robustness.

\paragraph{Keypoint Scoring:} 
For compute-constrained settings like edge devices, it is crucial to subsample keypoints to reduce memory and runtime in downstream tasks.
We claim that the inherent ranking of existing keypoint detectors via their confidence~\cite{superpoint,tyszkiewicz2020disk,zhao2023aliked} is suboptimal because it ignores the spatial distribution and matchability of points.
We propose to train a plug-and-play ranking head that maximizes the number of matches at different keypoint budgets and is compatible with all state-of-the-art keypoint detectors.

\paragraph{Spatial Uncertainty:}
Detections are subject to noise~\cite{germain2020s2dnet}, but their spatial covariance is rarely studied. 
However, estimating this is crucial for error propagation in downstream tasks, \eg, bundle adjustment in SfM~\cite{agarwal2010bundle,schoenberger2016sfm}.
Previous works quantified spatial uncertainty either through an up-to-scale anisotropic covariance~\cite{tirado2023dac,zeisl2009estimation} or a spatial confidence score~\cite{santellani2024gmm}. 
In contrast, we propose to learn a metric, anisotropic covariance estimator via homographic adaptation that can be used for end-to-end uncertainty propagation from keypoints to pose.

To summarize, in this work, we present 1) an isolated keypoint evaluation strategy that reflects modern requirements and challenges, 2) introduce a competitive keypoint detector, named \emph{\ours}, trained with reinforcement learning on synthetic homographies only, and finally 3) propose a simple yet effective strategy to estimate the keypoint ranking and its metric covariance without additional labels.

\section{Related Work}

\paragraph{Decoupling keypoints and descriptors} was predominant in the era of handcrafted local-features~\cite{lowe2004distinctive,bay2006surf,rublee2011orb}, which followed a \texttt{detect-then-describe} paradigm. There, interest points are first detected on corners~\cite{harris1988combined,shi1994good,lowe2004distinctive} and then robust descriptors are extracted for each keypoint~\cite{lowe2004distinctive}. Early works on learned local features adopted this approach~\cite{HardNet,barroso2019key,yi2016lift}. Later works proposed to couple both tasks under the \texttt{detect-and-describe} approach, where keypoints and descriptors are learned end-to-end in one network ~\cite{superpoint,dusmanu2019d2,revaud2019r2d2,zhao2022alike,zhao2023aliked}.
Recently, Li et al.~\cite{li2022decoupling} questioned this design decision and observed that weak descriptors can affect detection accuracy. 
Consequently, recent works decouple~\cite{li2022decoupling} both tasks, use completely separate networks for each~\cite{edstedt2024dedode,chen2025rdd,santellani2023strek,potje2024xfeat} or focus exclusively on detection~\cite{edstedt2025dad} or description~\cite{HardNet,wang2020learning,wang2023featurebooster}. Furthermore, recent success in feature matching~\cite{sarlin2020superglue,lindenberger2023lightglue,sun2021loftr,edstedt2023roma,edstedt2023dkm} reduced or alleviated the dependency on descriptors~\cite{edstedt2023roma,edstedt2023dkm}, but keypoints remain necessary for large-scale applications and classic SfM~\cite{schoenberger2016sfm}. Hence, we propose to train a lightweight detector independently and perform an unbiased evaluation of said detections using geometric correspondences.

\paragraph{Learned keypoint detection} has widely replaced and outperformed hand-crafted methods over the past years~\cite{Jin2020,imwchallenge2023}. 
A pioneering work is SuperPoint~\cite{superpoint}, which pre-trained on projections of synthetic shapes with clear corners, and finetuned with homography adaptation on random images to close the synthetic-to-real domain gap. 
Other works also couple keypoint detection and match success on more general image pairs~\cite{dusmanu2019d2,revaud2019r2d2,zhao2023aliked}, where MegaDepth~\cite{li2018megadepth} emerged as the de-facto standard training collection despite the lack of data diversity (phototourism). 
Some works address this limitation via differentiable pose estimation on posed images~\cite{bhowmik2020reinforced} or arbitrary image pairs~\cite{kunzel2025ripe}.
DeDoDe~\cite{edstedt2024dedode} learns tracks from large-scale SfM, while DISK~\cite{tyszkiewicz2020disk} used descriptors and ground truth depth or epipolar constraints to calculate reward values.
S-TREK~\cite{santellani2023strek} improves over DISK by replacing patch-based sampling, subject to border artifacts, with sequential sampling, and uses equivariant convolutions~\cite{weiler2019general,cesa2022program,lee2022self} to increase rotational robustness. 
In contrast, we show that state-of-the-art rotational stability can be achieved with effective data augmentations, and even without the added complexity and cost of such equivariant architectures.
RDD~\cite{chen2025rdd} and DeDoDe~\cite{edstedt2024dedode} decouple detection and description, while Edstedt et al.~\cite{edstedt2025dad} drop the dependence on descriptors in approaches trained with a policy gradient, and identify that light and dark detectors emerge. 
We also train our detector with a policy gradient, but on challenging homography adaptations of real images like in~\cite{christiansen2019unsuperpoint,wang2022rethinking}, but without any pretraining like SuperPoint~\cite{superpoint}.  

\paragraph{Keypoint uncertainty} is rarely studied in the literature. In practice, we are interested in uncertainty to i) filter keypoints which are unrepeatable, e.g. in the sky, ii) rank and sub-sample keypoints for increased computational efficiency and iii) propagating the spatial uncertainty to downstream algorithms.
Most networks output a keypoint score, which measures the model's confidence that a pixel is a keypoint~\cite{superpoint,tyszkiewicz2020disk,zhao2023aliked,revaud2019r2d2}.
This score often suffices to filter bad keypoints~\cite{zhao2023aliked}, but is suboptimal to rank keypoints because it ignores their spatial distribution and localization error, both important for accurate pose estimation~\cite{schoenberger2016sfm}.
Consequently, benchmarks often use a fixed number of keypoints~\cite{superpoint,zhao2023aliked,tyszkiewicz2020disk}, while keypoint subsampling, a critical hyperparameter in practice~\cite{sarlin2019coarse}, is rarely ablated~\cite{gleize2023silk}.
The spatial uncertainty of keypoints is studied by Muhle~\etal~\cite{muhle2023learning} using differentiable relative pose estimation.
DAC~\cite{tirado2023dac} proposes two post-hoc covariance estimates which are up-to-scale, derived from the score map of local feature extractors~\cite{zeisl2009estimation}.
UAPoint~\cite{zeng2025uncertainty} models aleatoric and epistemic uncertainty~\cite{kendall2017what} during training, which improves repeatability and matchability.
Santellani~\etal~\cite{santellani2024gmm} analyze the spatial variance of keypoints under different image augmentations in a gaussian mixture model (GMM) to both refine and score keypoints.
We study i) differentiable keypoint ranking to maximize repeatability after filtering, and ii) predicting metric, spatial covariance of our detections from homography adaptation, and their impact on downstream tasks.

%% file: figures/teaser/fig.tex
\begin{figure}[t]
    \centering
    \includegraphics[width=\linewidth]{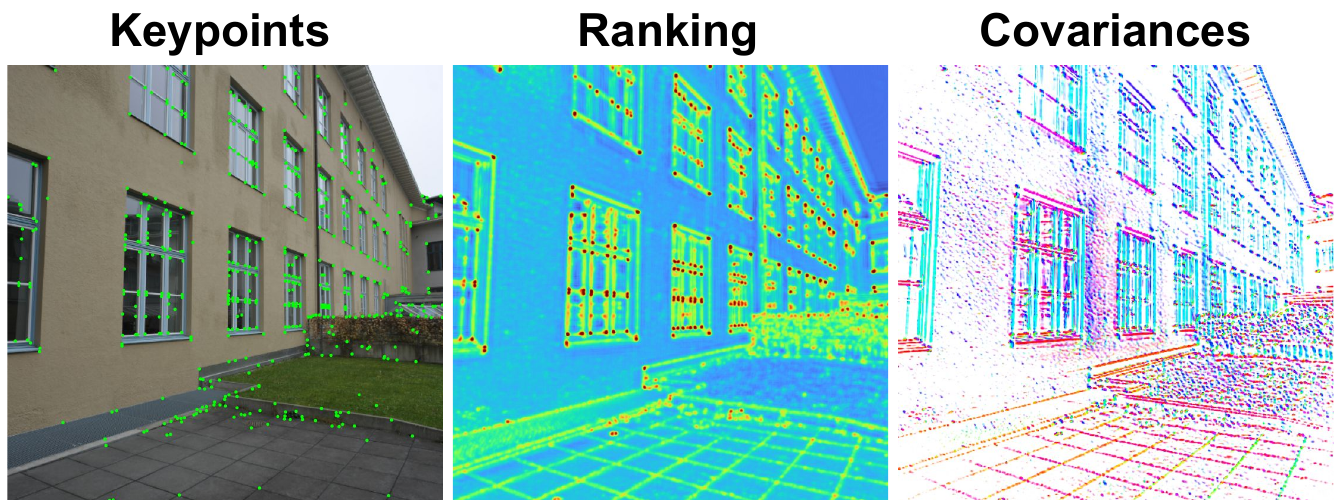}
    \caption{\textbf{Practical interest point detection.}
    \ours\ detects repeatable and interpretable corners (left), learned from perspective image crops. A dedicated ranking head (middle) maximizes the downstream accuracy-speed trade-off by ranking matchable points higher. The estimated 2D metric covariances (right) describe the keypoints' spatial uncertainty in pixels (colored by the angle of the first eigenvector, and whitened where the variance is large).}
    \label{fig:teaser}
\end{figure}

%% file: sec/5_method.tex
\section{Method}
\label{sec:method}

\begin{figure}[t]
  \centering
      \includegraphics[width=1.0\linewidth]{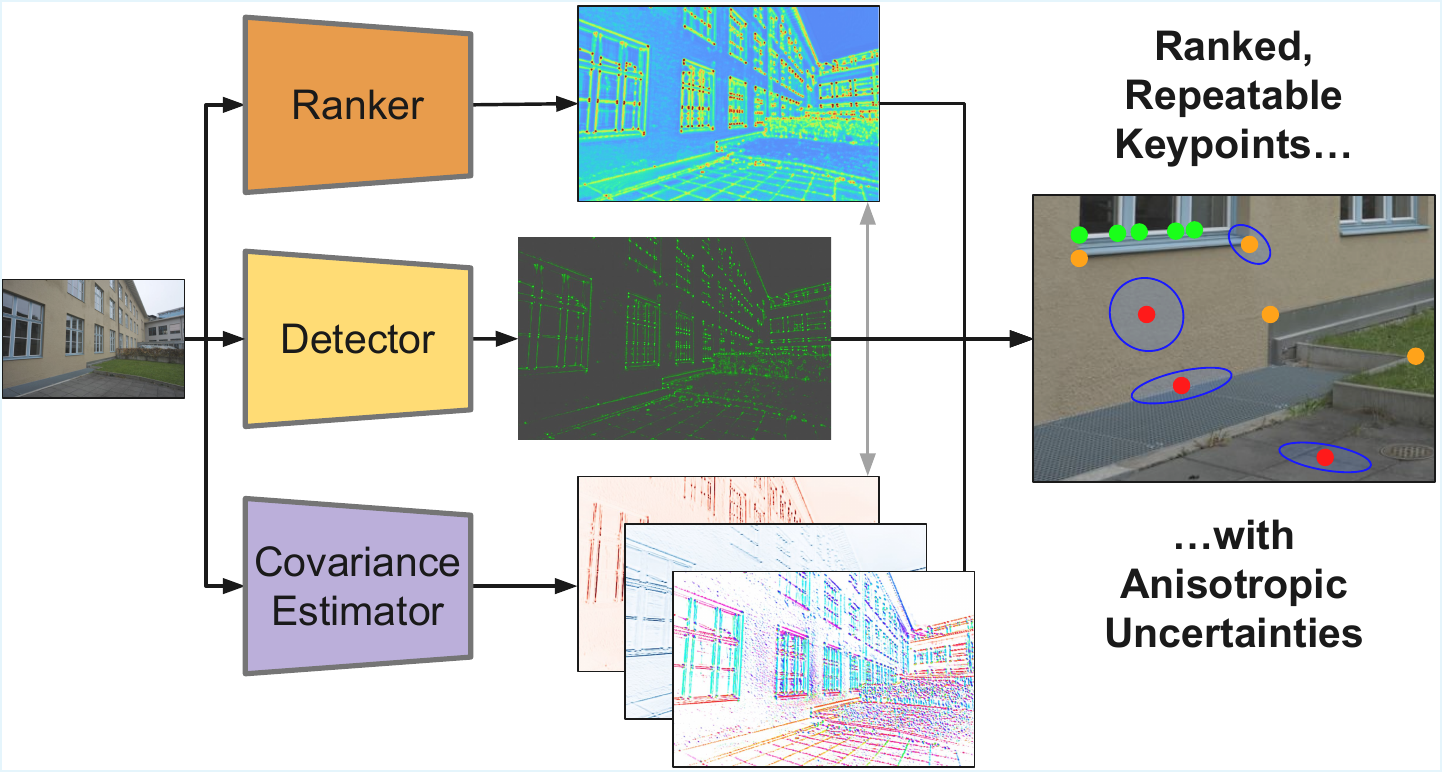}
      \caption{\textbf{Overview.} Our method consists of three branches: i) A detector head that produces a scoremap with repeatable keypoints, and ii) a covariance head that outputs the 2D spatial uncertainty in pixels, both sharing a lightweight backbone. The iii) ranker module outputs soft keypoint scores which maximize the repeatability at different keypoint budgets.
      }
      \label{fig:pipeline}
\end{figure}

We first provide an overview of our method, shown in ~\cref{fig:pipeline} and discussed in~\cref{ssec:overview}, illustrating our main components:
a \textbf{detector} (\cref{ssec:detector}) to find repeatable keypoints, a differentiable \textbf{ranker} (\cref{ssec:ranker}) to maximize repeatability at smaller keypoint budgets, and a \textbf{covariance estimator} (\cref{ssec:covariance}) to quantify their spatial uncertainty.

\subsection{Overview}
\label{ssec:overview}

\paragraph{Training Setup:}
\ours is trained on arbitrary image collections without any labels or auxiliary data.
In particular, we simulate two-view matching by sampling two crops from an image using synthetic homographies and strong photometric augmentations~\cite{superpoint,sarlin2020superglue}.

Let $\textbf{\textit{I}}_A, \textbf{\textit{I}}_B \in\mathbb{R}^{H{\times}W{\times}3}$ be two views of an image related by a known ground-truth homography $\*H_{A\rightarrow B}$.
The $i^{\text{th}}$ keypoint detected in view $v \in \{A, B\}$ is denoted by $\mathbf{x}_v^i$ or simply $\mathbf{x}^i$ and the set of detected keypoints by $\*x_v\in \mathbb{R}^{N\times 2}$.

\paragraph{Detector:} The detector module identifies keypoints that are repeatable, \ie, that can be reliably and accurately detected from multiple views and varying appearance conditions, usually located on corners or blobs.
Repeatability is a critical requirement for robust feature matching.
Given an input image $\textbf{\textit{I}}$, the detector estimates a heatmap $\textbf{\textit{P}}\in\mathbb{R}^{H\times W}$ that represents, for each pixel, the probability to be selected as keypoint.
We select keypoints $\*x_v$ as local maxima in this probability score map with non-maxima suppression (NMS), which prevents clustering of points~\cite{superpoint,zhao2023aliked,revaud2019r2d2,tyszkiewicz2020disk}.

Prior studies~\cite{santellani2023strek,lee2022self} have highlighted the lack of rotational equivariance in modern deep keypoint detectors and proposed architectural remedies, usually based on computationally expensive equivariant architectures~\cite{cesa2022program,weiler2019general}.
We instead demonstrate that extensive data augmentation during training is sufficient to effectively enforce rotation equivariance.
This approach enables our model to achieve strong robustness to image rotations while preserving a lightweight and efficient design.
Specifically, we generate training pairs using synthetic homographies with full $360^{\circ}$ rotations combined with strong photometric transformations.

\begin{figure}[t]
\centering
\includegraphics[width=1.0\linewidth]{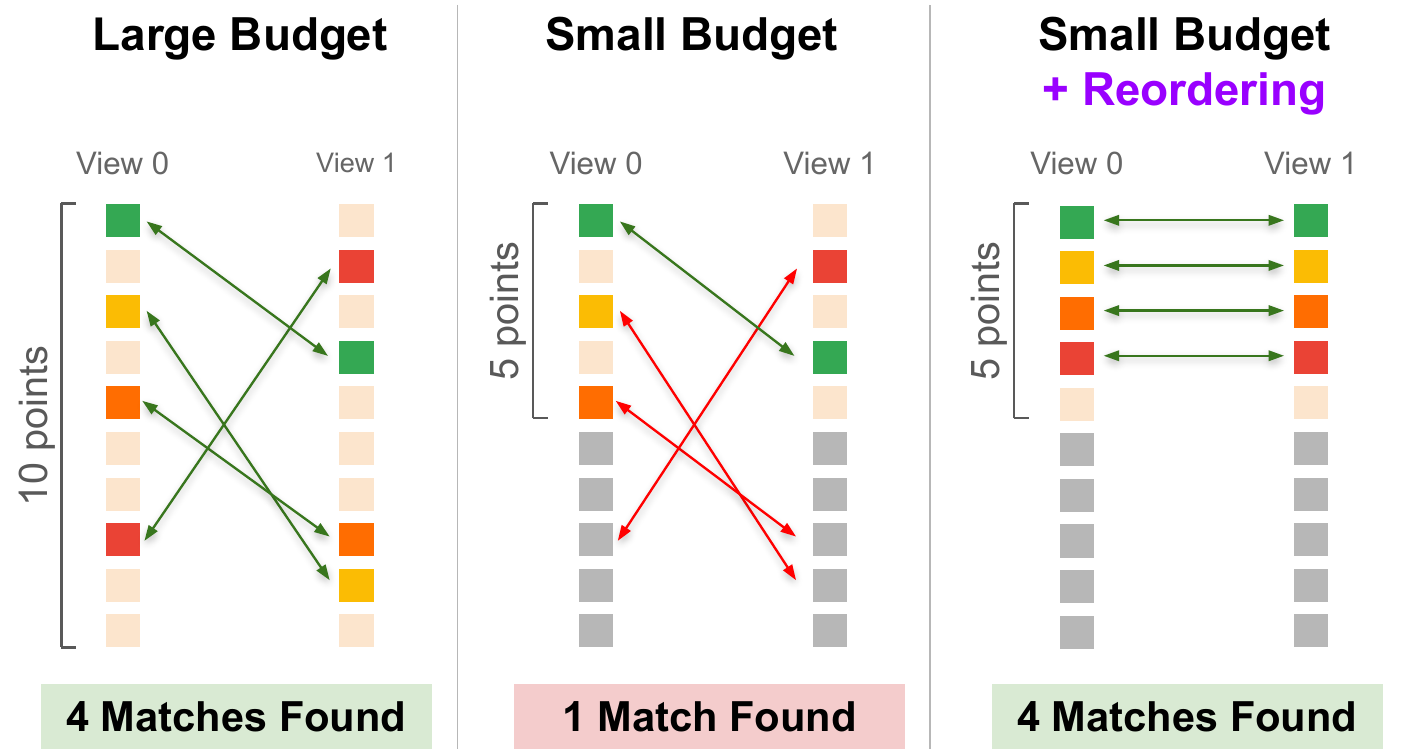}
\caption{\textbf{Keypoint ranking.} Inconsistent keypoint ranking between images (left) results in excessive match filtering when the amount of keypoints per image is restricted (\textit{small budget}, middle). Our ranking module keeps repeatable points at the top of the list and yields similar ranks for corresponding points (right).
}
\label{fig:ranker_setup}
\end{figure}

\paragraph{Ranker:} Modern, learned keypoint detectors typically order keypoints based on their detection score~\cite{superpoint,tyszkiewicz2020disk}, which indicates the likelihood of a keypoint's presence.
However, we want this ordering to maintain a maximum of matches, which requires awareness of the keypoints spatial distribution and matchability in each image.
Patch-based detectors~\cite{superpoint,tyszkiewicz2020disk}, in particular, tend to ignore this in their scores.
This can lead to a significant loss in matches and repeatability when the keypoint budget is limited, that is, the amount of sampled and retained keypoints $n\leq N$ is small.
\cref{fig:ranker_setup} demonstrates this problem: when considering fewer keypoints than trained for, a suboptimal ordering of the keypoints, where corresponding keypoints have vastly different ranks, leads to significantly fewer correspondences.

Our ranking module addresses this by providing an alternative ordering specifically designed to maximize matching performance over varying keypoint budgets.
This outputs a separate ranking score map $\textbf{\textit{R}} \in \mathbb{R}^{H \times W}$, from which we obtain a ranking score $r^i$ for each keypoint $\mathbf{x}^i$.

Our goal is to learn ranking scores for views $A$ and $B$, such that, when ordering the keypoints in descending order of ranking scores, the number of matches is maximized over \textit{all} keypoint budgets $n \in \{1, \dots, N\}$, where $N$ is the total number of detected keypoints. Let $\mathcal{R}^n(\*r_v) \in \mathbb{N}^{n}$ be the ordered set of $n$ keypoint indices in view $v$ with the highest ranking scores $\*r_v^i$.
The ranking problem can then be expressed as finding the optimal ranking scores $\*r_A$, $\*r_B$ that maximize
\begin{equation}
\label{eq:ranker_objective}
\max_{\*r_A, \*r_B} \sum_{n=1}^{N}|\textbf{M}(\mathcal{R}^n(\*r_A),\mathcal{R}^n(\*r_B))|\enspace.
\end{equation}
Here, $\textbf{M}: \mathbb{N}^n \times \mathbb{N}^n \to \mathbb{N}^{k\times2} \subseteq \mathcal{M}_{A\rightarrow B}$ extracts the subset of all ground-truth matches $\mathcal{M}_{A\rightarrow B}$ between the ranked and truncated keypoints in both views, and $|\textbf{M}|=k$:
\begin{equation}
\label{eq:matches_ranker}
\textbf{M}(\*x,\*y) = \{(i,j)\ \forall\; i,j \in \*x,\*y \text{ if } (i,j) \in \mathcal{M}_{A\rightarrow B}\}
\enspace
\end{equation}

\noindent \textbf{Covariance Estimator:} 
Accurately modeling the spatial uncertainty of detected keypoints is critical for robust 3D vision. While largely overlooked in recent years, propagating this uncertainty is necessary to obtain reliable covariance estimates for downstream tasks such as triangulation and pose estimation. Keypoints exhibit localization errors due to detector inaccuracies, image noise, and discretization effects, leading to reprojection errors across views. We characterize this error by estimating the 2D spatial uncertainty of keypoints in metric scale (pixels), providing a measure of confidence and local uniqueness of the keypoint.
Concretely, we want to estimate the symmetric, positive definite covariance matrix for each pixel, $\mathbf{\Sigma} \in \mathbb{R}^{H \times W \times 2 \times 2}$.

\subsection{Keypoint Detector}
\label{ssec:detector}

Following recent work in self-supervised keypoint learning~\cite{santellani2023strek, edstedt2025dad}, we adopt a \textit{policy-gradient}~\cite{NIPS1999_464d828b_policy_gradient} approach to train a detector that produces repeatable keypoints.

The detector takes a normalized RGB image $\textbf{\textit{I}}$ and outputs a keypoint score map $\textbf{\textit{S}} \in \mathbb{R}^{H \times W}$. This score map is normalized with a global softmax operation over the whole image to obtain a probability score map $\textbf{\textit{P}}\in\mathbb{R}^{H \times W}$. Keypoints are selected by applying Non-Maximum Suppression (NMS)~\cite{superpoint,zhao2023aliked,tyszkiewicz2020disk} followed by top-$N$ selection on this score map, which is a standard technique for keypoint sampling.

The training objective maximizes a reward signal which encourages repeatable keypoints to be detected in $\textbf{\textit{I}}_A$ and $\textbf{\textit{I}}_B$.

Following~\cite{santellani2023strek, edstedt2025dad}, the reward directly maximizes repeatability.
For a keypoint $\mathbf{x}_A^i$, it is defined as 
$\rho(\mathbf{x}_A^i) = \{\rho_\text{pos} \;\text{if } d(\mathbf{x}_A^i) \leq d_\text{max};\;\; \rho(\mathbf{x}_A^i) = \rho_\text{neg} \;\text{otherwise}\}$,
and $\rho_\text{pos}$ and $\rho_\text{neg}$ are the positive and negative reward, respectively. $d(\mathbf{x}_A^i)$ is the distance between the reprojected keypoint $\*H_{A\rightarrow B}(\mathbf{x}_A^i)$ and its closest neighbor in view $B$, and $d_\text{max}$ is a predefined radius for a successful match. 

We minimize the negative log-likelihood of the sampled keypoints, weighted by their normalized reward
\begin{equation}
\mathcal{L}_{\text{detector}} = -\sum_{v \in \{A, B\}}\sum_{i=1}^{K} \rho'(\mathbf{x}_v^i)\; \log p_v^i\enspace,
\end{equation}
where $\mathbf{x}_v^i$ is the $i$-th sampled keypoint from view $v \in \{A, B\}$, the term $p_v^i=\textbf{\textit{P}}_v[\mathbf{x}_v^i]$, where $[\cdot]$ is the lookup operator, and $p_v^i$ is the sampling probability of keypoint $\mathbf{x}_v^i$ according to the probability score map $\textbf{\textit{P}}_v$. The normalized reward $\rho(\mathbf{x}_v^i)$ is calculated following DaD~\cite{edstedt2025dad} as $\rho'(\mathbf{x}_v^i) = \frac{\rho(\mathbf{x}_v^i)}{\mathbb{E}_v[\rho(\mathbf{x_v^i})]+\epsilon}$, where $\rho(\mathbf{x}_v^i)$ is the un-normalized reward for that keypoint and $\mathbb{E}_v[\rho(\mathbf{x}_v^i)]$ is the average reward over all keypoints in view $v$.

\subsection{Ranker}
\label{ssec:ranker}

The discrete ranking objective in \cref{eq:ranker_objective} is non-differentiable. We use a differentiable approximation~\cite{blondel2020fast} to supervise our network via soft ranks. 
Let $h_{\mathrm{soft}}: \mathbb{R}^n \to \mathbb{R}^n$ denote the \emph{soft ranking} operator, which maps a vector of ranking scores $\mathbf{r} = [r^1, \dots, r^n]$ to their differentiable ranks $\mathbf{r}_{\mathrm{soft}} = h_{\mathrm{soft}}(\mathbf{r})$. We train the ranker module in a self-supervised manner using two loss terms. 

\vspace{0.5em}
\noindent \textbf{Spearman Loss:} 
Maximizing Spearman's rank correlation coefficient~\cite{Spearman1904} of our ordered keypoints encourages corresponding keypoints (matches) to have similar ranks within their respective lists. Maximizing this metric can be accomplished by minimizing the Euclidean distance between their soft ranks~\cite{blondel2020fast}. We compute the soft ranks for the subset of matched keypoints in both views, denoted by the vectors $\mathbf{r}_A^{\text{matched}}$ and $\mathbf{r}_B^{\text{matched}}$. 
The loss is then simply:
\begin{equation}
    \label{eq:rank_spearman_loss}
    \mathcal{L}_{\mathrm{spearman}} = 
    \frac{1}{N} \sum_{i=1}^{N} \left(h_{\text{soft}}(\mathbf{r}_{A, i}^{\text{matched}}) - h_{\text{soft}}(\mathbf{r}_{B, i}^{\text{matched}})\right)^2
\end{equation}

By minimizing this loss, we are encouraging that corresponding keypoints have similar ranks in both views.
This ensures that either both keypoints are present in the keypoint list or none are truncating.

\vspace{0.5em}
\noindent \textbf{Pull Loss:} 
In \cref{fig:ranker_setup} it is evident that the ranker needs to place the matched keypoints at the beginning of the list, and the unmatched points ranked at the end of the list. The \textit{pull} loss encourages this placement of keypoints by pulling the matched points towards the first rank ($\#1$) and the unmatched points towards the last rank ($\#N$). For each keypoint $\mathbf{x}_v^i$ with soft rank $h_{\mathrm{soft}}(r_v^i)$, the per-keypoint loss is:
\begin{equation}
    \label{eq:pull_rank_penalty}
    \mathcal{L}_{\text{pull}}^i = \begin{cases}
        |h_{\text{soft}}(r_v^i) - 1| & \text{if } \mathbf{x}_v^i \text{ is matched} \\
        |h_{\text{soft}}(r_v^i) - N| & \text{otherwise}
    \end{cases}
\end{equation}

The final ranker loss is a weighted combination of these two terms: $\mathcal{L}_{\text{ranker}} = \mathcal{L}_{\text{spearman}} + \lambda_{\text{ranker}} \cdot \frac{1}{N} \sum_{i=1}^{N} \mathcal{L}_{\text{pull}}^i$

\subsection{Covariance Estimator}
\label{ssec:covariance}

\begin{figure}[t]
\centering
\includegraphics[width=1.0\linewidth]{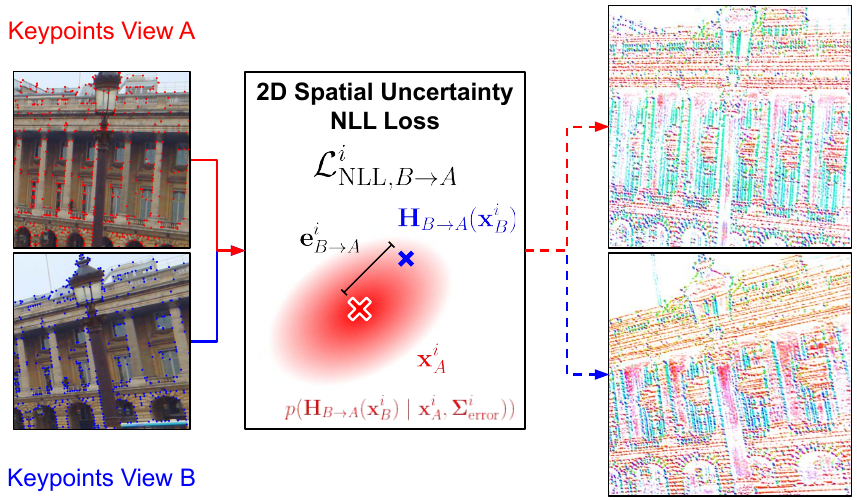}
\caption{\textbf{Covariance supervision.}
We train our covariance estimator by maximizing the log-likelihood of the reprojection error between corresponding keypoints. For corresponding keypoints $\mathbf{x}^i_A$ in view $A$ and $\mathbf{x}^i_B$ in view $B$, the reprojection error $\mathbf{e}^i_{B\rightarrow A} = \mathbf{H}_{B\rightarrow A}(\mathbf{x}^i_B) - \mathbf{x}^i_A$ is modeled as a zero-mean Gaussian: $\mathbf{e}^i_{B\rightarrow A} \sim \mathcal{N}(\mathbf{0}, \boldsymbol{\Sigma}^i_A + \mathbf{J}^i_{B\rightarrow A} \boldsymbol{\Sigma}^i_B (\mathbf{J}^i_{B\rightarrow A})^\top)$, where $\mathbf{J}^i_{B\rightarrow A}$ is the Jacobian of the homography evaluated at $\mathbf{x}^i_B$. The resulting covariance estimates (right) are strongly anisotropic (colored by the angle of the covariance's first eigenvector) and are large in areas with low texture (illustrated by opacity).
}
\label{fig:covariance_nll}
\end{figure}

We treat each detected keypoint $\mathbf{x}_v^i$ as a noisy observation with associated uncertainty $\boldsymbol{\Sigma}_v^i$. 
The reprojection error $\mathbf{e}_{B \rightarrow A}^i$ between corresponding keypoints accounts for uncertainties from both views:
$\mathbf{e}_{B \rightarrow A}^i = \mathbf{x}_A^i - \mathbf{H}_{B \rightarrow A}(\mathbf{x}_B^i) 
\sim \mathcal{N}(\mathbf{0}, \boldsymbol{\Sigma}_{\text{error}}^i)$,
where the combined error covariance is:
$\boldsymbol{\Sigma}_{\text{error}}^i = \boldsymbol{\Sigma}_A^i + \mathbf{J}_{B \rightarrow A}^i \boldsymbol{\Sigma}_B^i (\mathbf{J}_{B \rightarrow A}^i)^\top$,
and $\mathbf{J}_{B \rightarrow A}^i$ is the Jacobian of the homography transformation evaluated at $\mathbf{x}_B^i$, propagating the uncertainty from view B to view A.

Rather than predicting $\boldsymbol{\Sigma}_v^i$ directly, the network outputs the three non-zero elements of its Cholesky decomposition $\mathbf{L}_v^i$, where ${\boldsymbol{\Sigma}}_v^i = \mathbf{L}_v^i (\mathbf{L}_v^i)^\top$.
This parameterization guarantees symmetry and positive semi-definiteness. To ensure strict positive definiteness, the diagonal entries of $\mathbf{L}_v^i$ are passed through a Softplus activation. The network produces three scoremaps corresponding to the entries of $\mathbf{L}$, from which $\mathbf{L}_v^i$ is sampled at each detected keypoint location.

We compute the negative log-likelihood (NLL) of the reprojection error as
\begin{equation}
\label{eq:NLL_combined}
\begin{split}
\mathcal{L}^i_{\text{NLL}, B \rightarrow A} = & \frac{1}{2}\log \det (\boldsymbol{\Sigma}_{\text{error}}^i) \\
& + \frac{1}{2} (\mathbf{e}_{B \rightarrow A}^i)^\top (\boldsymbol{\Sigma}_{\text{error}}^i)^{-1} \mathbf{e}_{B \rightarrow A}^i
\enspace,
\end{split}
\end{equation}
where the matrix inverse and determinant are computed efficiently via Cholesky decomposition of $\boldsymbol{\Sigma}_{\text{error}}^i$.
We also apply the same loss in the reverse direction ($A \rightarrow B$) and define the total covariance loss as the average bidirectional NLL over all $|\bf{M}|$ matched keypoints:
\begin{equation}
\label{eq:covariance_loss}
\mathcal{L}_{\text{covariance}} = \frac{1}{2|\bf{M}|} \sum_{i=1}^{|\bf{M}|}
\left(\mathcal{L}^i_{\text{NLL}, B \rightarrow A} +
    \mathcal{L}^i_{\text{NLL}, A \rightarrow B}\right).
\end{equation}

%% file: sec/6_experiments_results.tex
\section{Experiments}
\label{sec:experiments}

We first report important implementation details in~\cref{ssec:implementation}. 
In~\cref{ssec:det_matching}, we study the detected keypoints on four two-view estimation benchmarks, followed by detailed evaluations on rotational equivariance in~\cref{ssec:rotationequivariance}.
Finally, we ablate the proposed keypoint ranking and covariance estimation modules in~\cref{ssec:rankingeval} and~\cref{ssec:coveval}, respectively.

\subsection{Implementation}
\label{ssec:implementation}
The architecture of RaCo is based on that of ALIKED-N(16)~\cite{zhao2023aliked}, which is lightweight, multi-scale, and sufficiently expressive.
It predicts both the detection heatmap and the pixelwise covariance.
The ranker is a separate ResNet~\cite{he2016deep} backbone which takes as input the normalized RGB image and outputs the ranker score map $\textbf{\textit{R}}$.
We first train the backbone and detection head, and later train the covariance head and the ranker separately in a second stage.
The model is trained on synthetic pairs sampled from the Oxford-Paris 1M distractors dataset~\cite{radenovic2018revisiting}.
It is trained using the AdamW~\cite{loshchilov2017Fixing} optimizer in PyTorch~\cite{pytorch} with an NVIDIA 2080 Ti GPU.
The detector head is trained using 400k samples center cropped at $768\times 768$ and then resized to $640\times 640$ after sampling the homographies, for 1 epoch with a batch size of 2 with an initial learning rate of $2\times10^{-4}$ decaying according to a cosine annealing schedule to a terminal learning rate of $10^{-6}$.
We sample 512 keypoints per image with an NMS radius of 3px following the training sampling of~\cite{edstedt2025dad}.

For the detector we set $d_{\text{max}}{=}$1.2px, $\rho_\text{pos}{=}$1, and $\rho_\text{neg}{=}-\min\{10^{-2}, t\cdot10^{-6}\}$ where $t$ is the number of optimizer steps taken through training. At inference time we use subpixel sampling based on the soft-argmax over the patch around the selected keypoint~\cite{edstedt2025dad, zhao2023aliked}.

For the training of the other two modules, the inference setting of the detector is used to capture the distribution of keypoints at inference. 
The covariance estimator head is trained for 20k steps and the ranker module is separately trained for 1 epoch with the identical setup as the detector training. 
For both training runs we freeze all model layers except those of the pertinent module/head. More details are provided in the supplementary.

\subsection{Two-View Keypoint Matching}
\label{ssec:det_matching}
\paragraph{Setup:} We present a comprehensive evaluation of our keypoint detector in the two-view setting. We use 4 different evaluation datasets and use the ground truth transformations to project keypoints across views. Keypoint matching is subsequently performed by identifying mutual nearest neighbors within a specified reprojection radius in both views. We force all detectors to detect the same number of keypoints.

HPatches~\cite{balntas2017hpatches} comprises over 500 real-world image pairs under a homography transformation. The views are subject to either illumination or viewpoint changes. 
DNIM~\cite{zhou2016evaluating} consists of 1722 images grouped into 17 sequences per webcam. This data set exhibits strong illumination changes, and we sample 428 random image pairs augmented with random homographies to also evaluate the robustness to perspective changes.
MegaDepth~\cite{li2018megadepth} is a large-scale dataset of photo-tourism internet images.
We use the subset MegaDepth1800~\cite{lindenberger2023lightglue} of 4 scenes from the dataset's test set.
We introduce the ETH3D-Two-View dataset, covisible image pairs sampled from indoor and outdoor scenes in ETH3D~\cite{schops2017multi}.
ETH3D and MegaDepth provide ground truth camera poses, intrinsic parameters, and depth images, using which we can map points across images.

\paragraph{Baselines:} We compare our model against SIFT~\cite{lowe2004distinctive} and several learned keypoint detectors.
SuperPoint~\cite{superpoint} is trained on homographies in a supervised way to explicitly detect corners. 
ALIKED-N(16)~\cite{zhao2023aliked} is trained in an unsupervised manner using depth and homography data.
DISK~\cite{tyszkiewicz2020disk} and DaD~\cite{edstedt2025dad} are both trained on depth data.

We report the number of ground-truth matches within a reprojection threshold, the fraction of repeatable points within two thresholds and the localization error in pixels.
We also estimate homographies ($\mathbf{H}$) or relative poses ($\mathbf{T}$) and report the Area Under the recall Curve (AUC).

\paragraph{Results:} \cref{tab:hpatches} and \cref{tab:multi_dataset_pose} summarize the results of the matching evaluation on both types of datasets. On all datasets, our model obtains the highest repetability at 3px. 
Our model is competitive with ALIKED~\cite{zhao2023aliked} and DaD~\cite{edstedt2025dad}, which are trained with depth supervision, on relative pose estimation and our localization error is slightly higher than DaD~\cite{edstedt2025dad} and equal to ALIKED~\cite{zhao2023aliked} on ETH3D.

DISK~\cite{tyszkiewicz2020disk}, which is trained on MegaDepth~\cite{li2018megadepth}, exhibits strong performance in terms of number of matches, however, at the cost of repeatability and weaker generalization to ETH3D or HPatches, which have a slight domain gap compared to MegaDepth. SuperPoint~\cite{superpoint}, despite not having competitive repeatability scores, is still able to estimate the relative pose very well, which shows that corners are high quality keypoints and that depth data is not necessarily required to train a good detector. On DNIM~\cite{zhou2016evaluating}, \ours\ outperforms all other methods, showing the model's robustness to both illumination and perspective changes.

\begin{table}[ht]
\centering
\small
\setlength{\tabcolsep}{2.99pt}
\begin{tabular}{llcccccc}
\toprule
& \multirow{2}{*}[-0.3em]{detector}
& \multirow{2}{*}[-0.3em]{\makecell{$\#$matches\\@3px}} 
& \multicolumn{2}{c}{{rep.\ [\%]}} 
& \multirow{2}{*}[-0.3em]{\makecell{loc.\\{[px]}}} 
& \multicolumn{2}{c}{{AUC $\mathbf{H}$}} \\
\cmidrule(lr){4-5}
\cmidrule(lr){7-8}
& & & 1px & 3px & & 1px & 3px \\
\midrule
\multirow{6}{*}{\rotatebox{90}{HPatches~\cite{balntas2017hpatches}}} 
& SIFT~\cite{liu2010sift} & 282 & 24.8 & 44.0 & 0.99 & 32.7 & 68.8 \\
& SuperPoint~\cite{superpoint} & 503 & 22.1 & 54.0 & 1.19 & \cellcolor{tabfirst}41.5 & \cellcolor{tabfirst}74.7 \\
& DISK~\cite{tyszkiewicz2020disk} & 484 & 23.9 & 49.9 & 1.05 & 27.5 & 63.7 \\
& ALIKED~\cite{zhao2023aliked} & \cellcolor{tabsecond}530 & \cellcolor{tabsecond}32.8 & \cellcolor{tabsecond}56.3 & \cellcolor{tabfirst}0.96 & 34.1 & 69.5 \\
& DaD~\cite{edstedt2025dad} & 510 & 30.2 & 55.2 & 1.01 & 40.0 & 73.5 \\
& Ours & \cellcolor{tabfirst}544 & \cellcolor{tabfirst}33.9 & \cellcolor{tabfirst}58.5 & \cellcolor{tabsecond}0.98 & \cellcolor{tabsecond}41.2 & \cellcolor{tabsecond}74.2 \\
\midrule
\multirow{4}{*}{\rotatebox{90}{DNIM~\cite{zhou2016evaluating}}} 
& SuperPoint~\cite{superpoint} & \cellcolor{tabsecond}56 & 11.9 & \cellcolor{tabsecond}30.2 & 1.26 & \cellcolor{tabsecond}2.4 & \cellcolor{tabsecond}23.1 \\
& ALIKED~\cite{zhao2023aliked} & 57 & 11.6 & 26.6 & \cellcolor{tabsecond}1.13 & 1.7 & 18.0 \\
& DaD~\cite{edstedt2025dad} & 58 & \cellcolor{tabsecond}13.9 & 29.8 & 1.16 & 1.9 & 23.2 \\
& Ours & \cellcolor{tabfirst}72 & \cellcolor{tabfirst}20.4 & \cellcolor{tabfirst}35.8 & \cellcolor{tabfirst}1.00 & \cellcolor{tabfirst}4.4 & \cellcolor{tabfirst}25.9 \\
\bottomrule
\end{tabular}
\caption{\textbf{Homography estimation.} We evaluate the repeatability and correspondences of keypoints paired with ground-truth matches on HPatches~\cite{balntas2017hpatches} and DNIM~\cite{zhou2016evaluating}. 
Our method achieves competitive repeatability and pose estimation performance on HPatches, and exhibits superior robustness on image pairs from DNIM with large illumination changes. We color the \colorbox{tabfirst}{best} and \colorbox{tabsecond}{second best} results for each metric are colored.
}
\label{tab:hpatches}
\end{table}

\begin{table}[ht]
\centering
\small
\setlength{\tabcolsep}{2.9pt}
\begin{tabular}{llcccccc}
\toprule
& \multirow{2}{*}[-0.3em]{detector}
& \multirow{2}{*}[-0.3em]{\makecell{$\#$matches\\@3px}} 
& \multicolumn{2}{c}{rep.\ [\%]} 
& \multirow{2}{*}[-0.3em]{\makecell{loc.\\{[px]}}} 
& \multicolumn{2}{c}{AUC $\mathbf{T}$} \\
\cmidrule(lr){4-5}
\cmidrule(lr){7-8}
&  & & 3px & 5px & & {5\degree} & {10\degree} \\
 \midrule
\multirow{6}{*}{\rotatebox[origin=c]{90}{MD1800~\cite{lindenberger2023lightglue}}} & SIFT~\cite{liu2010sift} & 317 & 34.8 & 39.0 & 1.47 & 66.0 & 78.7 \\
 & SuperPoint~\cite{superpoint} & 549 & 42.5 & 46.8 & 1.61 & 71.1 & 82.3 \\
 & DISK~\cite{tyszkiewicz2020disk} & \cellcolor{tabfirst}647 & 41.0 & 43.7 & \cellcolor{tabsecond}1.39 & 67.9 & 80.2 \\
 & ALIKED~\cite{zhao2023aliked} & 579 & \cellcolor{tabsecond}45.9 & \cellcolor{tabsecond}48.6 & \cellcolor{tabfirst}1.30 & 71.2 & 82.5 \\
 & DaD~\cite{edstedt2025dad} & 588 & \cellcolor{tabsecond}45.9 & \cellcolor{tabfirst}49.9 & 1.51 & \cellcolor{tabfirst}72.4 & \cellcolor{tabfirst}83.3 \\
 & Ours & \cellcolor{tabsecond}595 & \cellcolor{tabfirst}46.1 & \cellcolor{tabfirst}49.9 & 1.44 & \cellcolor{tabsecond}71.8 & \cellcolor{tabsecond}82.8 \\
\midrule
\multirow{6}{*}{\rotatebox[origin=c]{90}{ETH3D~\cite{schops2017multi}}} & SIFT~\cite{liu2010sift} & 298 & 34.6 & 41.5 & 1.62 & 85.2 & 90.2 \\
 & SuperPoint~\cite{superpoint} & 498 & 42.6 & 47.0 & 1.57 & \cellcolor{tabsecond}92.9 & \cellcolor{tabsecond}96.1 \\
 & DISK~\cite{tyszkiewicz2020disk} & \cellcolor{tabsecond}559 & 43.1 & 48.2 & 1.50 & 85.0 & 89.6 \\
 & ALIKED~\cite{zhao2023aliked} & 523 & 43.9 & 47.7 & 1.38 & 90.3 & 94.0 \\
 & DaD~\cite{edstedt2025dad} & 534 & \cellcolor{tabsecond}46.5 & \cellcolor{tabsecond}50.1 & \cellcolor{tabfirst}1.21 & \cellcolor{tabfirst}94.5 & \cellcolor{tabfirst}96.8 \\
 & Ours & \cellcolor{tabfirst}562 & \cellcolor{tabfirst}47.2 & \cellcolor{tabfirst}52.7 & \cellcolor{tabsecond}1.37 & 92.5 & 95.6 \\
\bottomrule
\end{tabular}
\caption{\textbf{Relative pose estimation.}
We evaluate two-view matching performance and keypoint repeatability on two datasets:
MegaDepth1800~\cite{lindenberger2023lightglue,li2018megadepth} and ETH3D-Two-View~\cite{schops2017multi}.
The colors indicate the \colorbox{tabfirst}{best} and \colorbox{tabsecond}{second best} results for each metric.
On both benchmarks, our method achieves the highest repeatability despite only being trained on homographies.
}\label{tab:multi_dataset_pose}
\end{table}

\subsection{Rotation Equivariance}
\label{ssec:rotationequivariance}
\paragraph{Setup:} Following~\cite{rublee2011orb}, we evaluate the rotation equivariance of keypoint detectors using in-plane rotations. We take the first 20 images from HPatches~\cite{balntas2017hpatches} and take the largest square crop that can be rotated through $360^\circ$. We rotate this original view in increments of ${10}^\circ$, and resize it to $512\times 512$. We add zero-mean Gaussian noise ($\sigma=10$, pixel intensity scale $[0,255]$) to each image pair to suppress interpolation artifacts. We run the keypoint detectors on this image pair and extract 200 keypoints per view.

\paragraph{Baselines:}
We evaluate ALIKED-N(16)~\cite{zhao2023aliked} (trained with and without extra rotation augmentation), DaD~\cite{edstedt2025dad}, SuperPoint~\cite{superpoint}, DISK~\cite{tyszkiewicz2020disk}, SIFT~\cite{lowe2004distinctive}, and our model.
We report the repeatability at thresholds of $\{1,2,3\}$ px.
We summarize the rotation equivariance as the area under the repeatability vs rotation angle curve (rotation AUC) after normalizing the angles to the range $[0,1]$.
We also report the average runtime of each detector.

\paragraph{Results:} 

\begin{figure}[t] 
    \centering
    \includegraphics[width=0.9\linewidth]{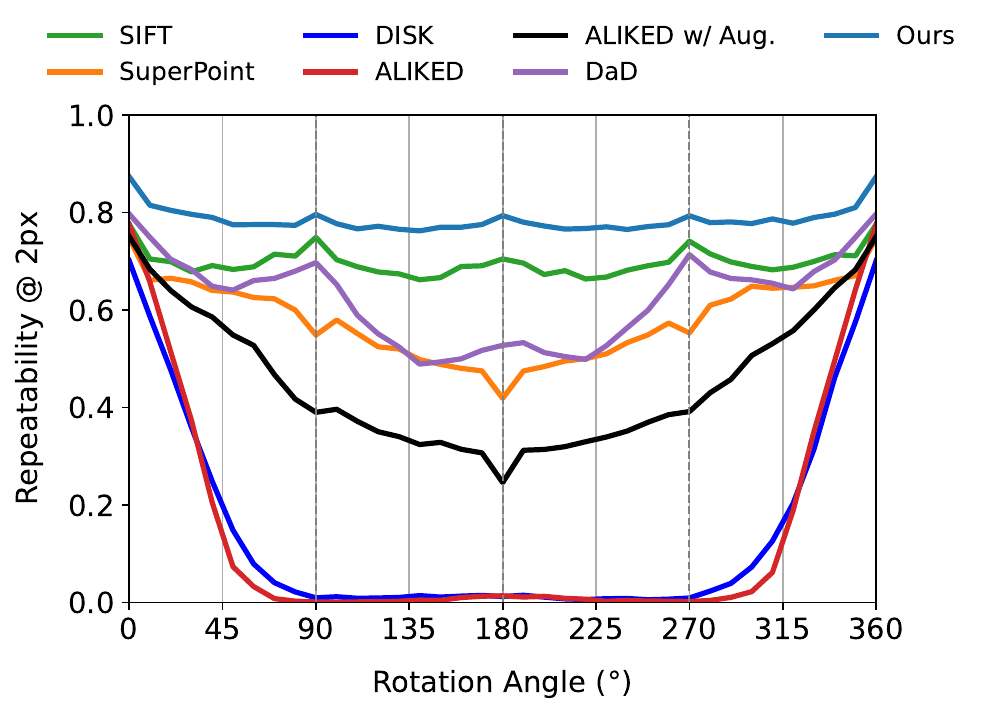}
    \vspace{-0.5em}%
    \caption{\textbf{Rotation evaluation on HPatches~\cite{balntas2017hpatches}.} We plot the repeatability@2px over the rotation angle between image pairs. SIFT~\cite{lowe2004distinctive} is more robust than any learned keypoint detectors, but our improved rotation augmentations result in state-of-the-art rotational robustness without requiring specialized model architectures.}%
    \label{fig:rotation_eval}
\end{figure}

\cref{fig:rotation_eval} illustrates the repeatability at various rotation angles and \cref{tab:rotation_auc} reports the AUC of the repeatability up to different thresholds.
Our model retains the highest repeatability by a large margin when the image pair is subject to in plane rotation.
We achieve a consistent repeatability around 80\% over all angles while other detectors with the exception of SIFT exhibit a degradation in repeatability. 

In contrast to DaD~\cite{edstedt2025dad}, which uses rotation augmentations at 90\textdegree\ intervals, we sample rotations over the entire circle, resulting in more stable and generally increased equivariance.
Training \ours\ with rotationally equivariant convolutions~\cite{cesa2022program} results in $3\%$ increased repeatability, but is $10{\times}$ slower at inference and $3.5{\times}$ slower to train, consuming $2.5{\times}$ more memory.
Removing rotation augmentations drastically impairs the accuracy of the models, demonstrating the importance of smooth rotation augmentations.

\begin{table}
    \centering
    \small
    \setlength{\tabcolsep}{4pt} %
    \begin{tabular}{l*{3}{>{\centering}p{1cm}}c}
    \toprule
     \multirow{2}{*}[-0.3em]{detector}
     & \multicolumn{3}{c}{repeatability\ AUC [$\%$]} 
     & \multirow{2}{*}[-0.3em]{\makecell[c]{run time \\{[ms]}}} \\
    \cmidrule(lr){2-4}
     & 1px & 2px & 3px & \\
    \midrule
    SIFT~\cite{lowe2004distinctive} & \cellcolor{tabsecond}52.3 & \cellcolor{tabsecond}69.6 & \cellcolor{tabsecond}71.9 & 34.8 \\
    SuperPoint~\cite{superpoint} & 25.3 & 57.7 & 67.7 & \06.8 \\
    DISK~\cite{tyszkiewicz2020disk} & 7.3 & 13.0 & 18.1 & 31.6 \\
    ALIKED~\cite{zhao2023aliked} & 8.4 & 12.8 & 16.7 & \cellcolor{tabsecond}\05.1 \\
    ALIKED w/ Aug.~\cite{zhao2023aliked} & 24.9 & 44.8 & 49.7 & \cellcolor{tabsecond}\05.1\\
    DaD~\cite{edstedt2025dad} & 27.1 & 62.0 & 70.0 & 15.2 \\
    Ours & \cellcolor{tabfirst}71.5 & \cellcolor{tabfirst}78.3 & \cellcolor{tabfirst}79.5 & \cellcolor{tabfirst}4.8 \\
    \midrule
    Ours w/ ReCONV\cite{cesa2022program} & \cellcolor{tabfirst}74.8 & \cellcolor{tabfirst}81.9 & \cellcolor{tabfirst}83.1 & 42.3 \\
    Ours w/o Aug. & 13.8 & 24.9 & 33.1 & \cellcolor{tabfirst}4.8 \\
    \bottomrule
    \end{tabular}
    \caption{\textbf{Rotation ablation and run time.}
    We report the area under the repeatability curve up to different error thresholds under in-plane rotations up to 360\textdegree.
    The upper section compares our model against state-of-the-art keypoint detectors and the lower section ablates key design decisions like rotation augmentations and equivariant convolutions.
    We color the \colorbox{tabfirst}{best} and \colorbox{tabsecond}{second best} results for each metric.
    Our method significantly outperforms state-of-the-art models because of strong rotation augmentations, even without expensive equivariant convolutions.
    }%
    \label{tab:rotation_auc}
\end{table}

\subsection{Keypoint Ranking}
\label{ssec:rankingeval}

\paragraph{Setup:} We consider the setup of the matching evaluation on HPatches~\cite{balntas2017hpatches} and MegaDepth1800~\cite{lindenberger2023lightglue} from \cref{ssec:det_matching}. We vary the number of keypoints extracted per view (keypoint budget), and compute the repeatability at that budget. We provide more details in the supplementary material.

\paragraph{Baselines:} We compare our model in two settings, the first when the keypoint probability scores from the detector score map $\textbf{\textit{P}}$ are used to order the points, and secondly when the ranking scores from $\textbf{\textit{R}}$ are used to order  (\textit{+Ranker}).
To demonstrate that our ranker model is applicable to any detector, we retrain the ranker module with SuperPoint's keypoints and include it in the comparison. 

\begin{figure}[t]
    \centering
    \begin{subfigure}{0.93\linewidth}
        \centering
        \includegraphics[width=\linewidth]{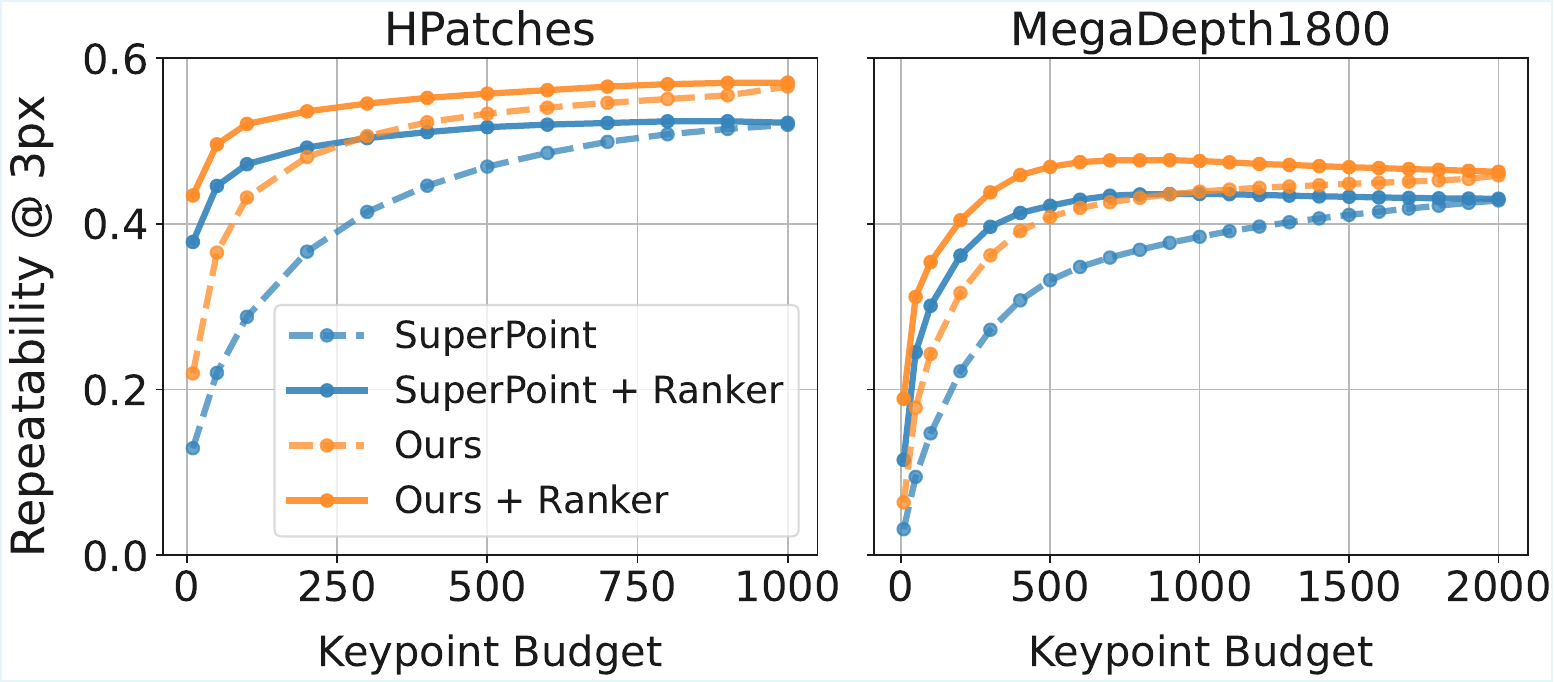}

        \label{fig:ranker_repeatability}
    \end{subfigure}

    \begin{subfigure}{0.95\linewidth}
        \centering
        \includegraphics[width=\linewidth]{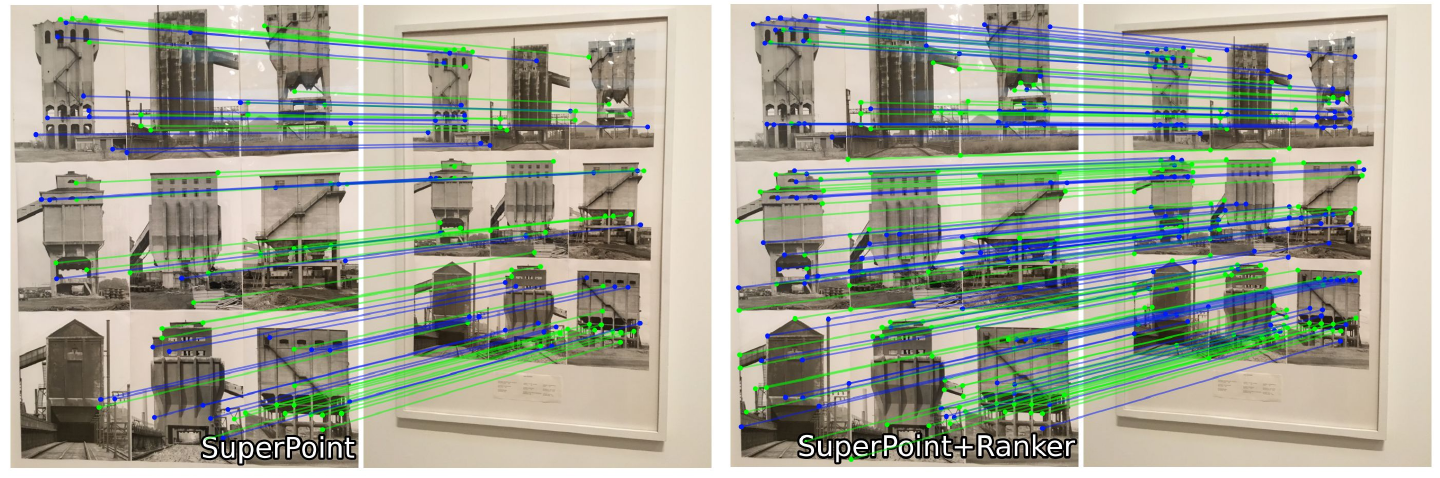}
        \label{fig:superpoint_ranker}
    \end{subfigure}

    \vspace{-1.0em}
    
    \caption{\textbf{Keypoint ranking evaluation.} 
    We evaluate the repeatability at 3px against the number of keypoints per image on HPatches~\cite{balntas2017hpatches} and MegaDepth1800~\cite{lindenberger2023lightglue} (top).
    For both SuperPoint~\cite{superpoint} and \ours, our ranker module is more effective in ordering keypoints than the original detection scores.
    We also visualize the repeatable points as matches at a budget of 128~(${\color{green}\bullet}$) and 256~(${\color{blue}\bullet}$) keypoints (bottom). 
    SuperPoint+Ranker (right) almost doubles repeatable points here over SuperPoint~\cite{superpoint} scores (bottom left). 
    }
    \label{fig:ranker_combined}
\end{figure}

\paragraph{Results:} For both our model and SuperPoint, ordering and truncating keypoints according to our ranking score provides a significant boost to the repeatability at all keypoint budgets on both datasets, see \cref{fig:ranker_combined}. Note how the terminal repeatability is identical for models with and without the ranker, because it is only performing a reordering operation.
We observed that keypoint detectors that select keypoints over a grid such as SuperPoint~\cite{superpoint} or DISK~\cite{tyszkiewicz2020disk} suffer more from a bad ordering of points. 
Especially there, our ranker is able to recover repeatability at restricted budgets by ranking points over the entire image.
Keypoint detectors that produce globally normalized heatmaps~\cite{santellani2023strek, edstedt2025dad, zhao2023aliked} (also our method) are typically trained with some sort of top-K selection, and hence, an implicit global ordering. At train time, they learn to prioritize matchable keypoints, similar to our ranker.
However, they do not enforce the correlation between matched points. This being a different objective than detection probabilities motivates our dedicated ranker module, and results in further improved truncation robustness.

\subsection{Multiview Triangulation}
\label{ssec:coveval}
\paragraph{Setup:} We evaluate the covariances for the task of 3D triangulation on the ETH3D dataset~\cite{schops2017multi}.
We reproject keypoints using the ground truth depth and poses across views to form matches, and subsequently triangulate them to multi-view tracks~\cite{schoenberger2016sfm,sarlin2019coarse}. 
We then refine the 3D points with non-linear least squares optimization, implemented with PyCeres and COLMAP~\cite{ceres-solver,schoenberger2016sfm}, in which reprojection error residuals are weighted by the estimated covariances.
Post-convergence, we compute 3D marginal covariances, defining precision as the reciprocal of the covariance ellipsoid volume. 
Points are sorted by precision and pruned from lowest to highest to reach the target size. This filters for high triangulation stability relative to input 2D uncertainties.

\paragraph{Baselines:} We compare our metric 2D spatial uncertainties against the following baselines. Detector Agnostic Covariances: DAC~\cite{tirado2023dac} provides isometric (\textbf{DAC-iso}) and full (\textbf{DAC-full}) covariances up to scale by operating on the detector score maps.
The next baseline involves assigning a \textbf{Constant} isotropic covariance to every keypoint in the image.
Finally we weight the keypoints by the reprojection error (\textbf{Reproj. Err.}) from the 3D reconstruction.

\begin{figure}[t]
    \centering
    \includegraphics[width=0.93\linewidth]{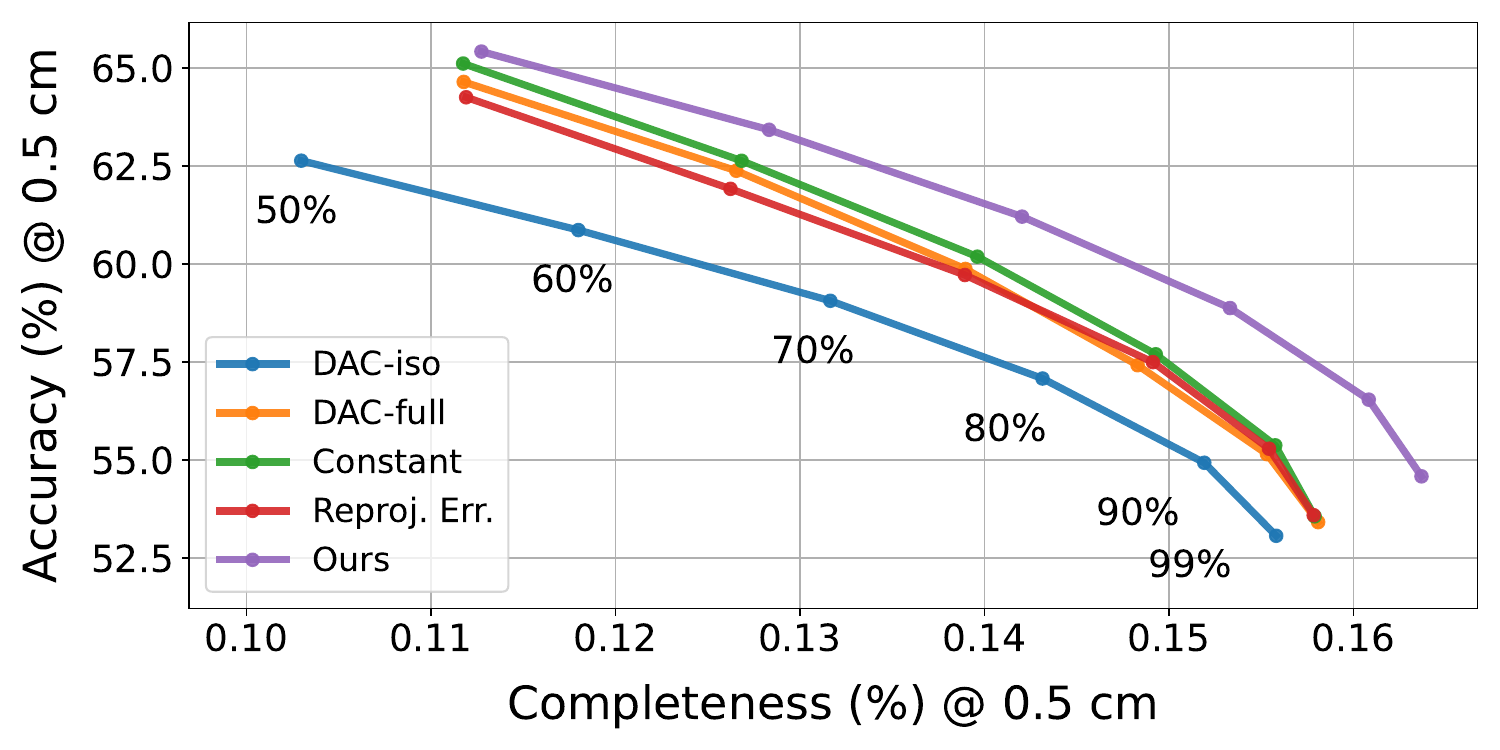}

    \caption{
    \textbf{3D triangulation on ETH3D~\cite{schops2017multi}.} We plot triangulated point cloud accuracy against completeness (within 0.5cm). Each point represents the fraction of the original cloud retained after filtering by spatial uncertainty metrics. Our 2D covariances i) improve both accuracy and completeness even without filtering, and ii) yield superior 3D covariance estimates for more effective filtering.
    }
    \label{fig:filtering_acc_vs_completness}
\end{figure}

\paragraph{Results:} \cref{fig:filtering_acc_vs_completness} shows that across all point cloud sizes, our metric 2D covariances consistently attains the highest accuracy and completeness. 
Interestingly, assuming a \textbf{Constant} isotropic keypoint covariance remains surprisingly competitive, likely because of large track lengths.
\textbf{DAC-full}'s anisotropic covariances perform better than \textbf{DAC-iso}, but the lack of scale in both methods limit the informativeness of their covariances in this multi-view setting.
Our method achieves highest performance even at $99\%$, suggesting that our estimated 2D covariances also increase the accuracy of 3D points by downweighting noisy observations.

\subsection{Covariance Metric Consistency}
\label{ssec:covcalib}

\paragraph{Setup:} Following \cref{ssec:coveval}, we bin points by the square root of the trace of their predicted marginal covariances into 20 equally populated bins. We plot the mean predicted uncertainty against the mean observed Euclidean distance to the ground truth mesh for each bin.

\paragraph{Baselines:} We evaluate the same baselines as in \cref{ssec:coveval}. We compute the slope $\beta$ via log-log linear regression, where a slope of $\beta=1$ indicates perfect metric consistency.

\begin{figure}[t] 
    \centering
    \includegraphics[width=0.85\linewidth]{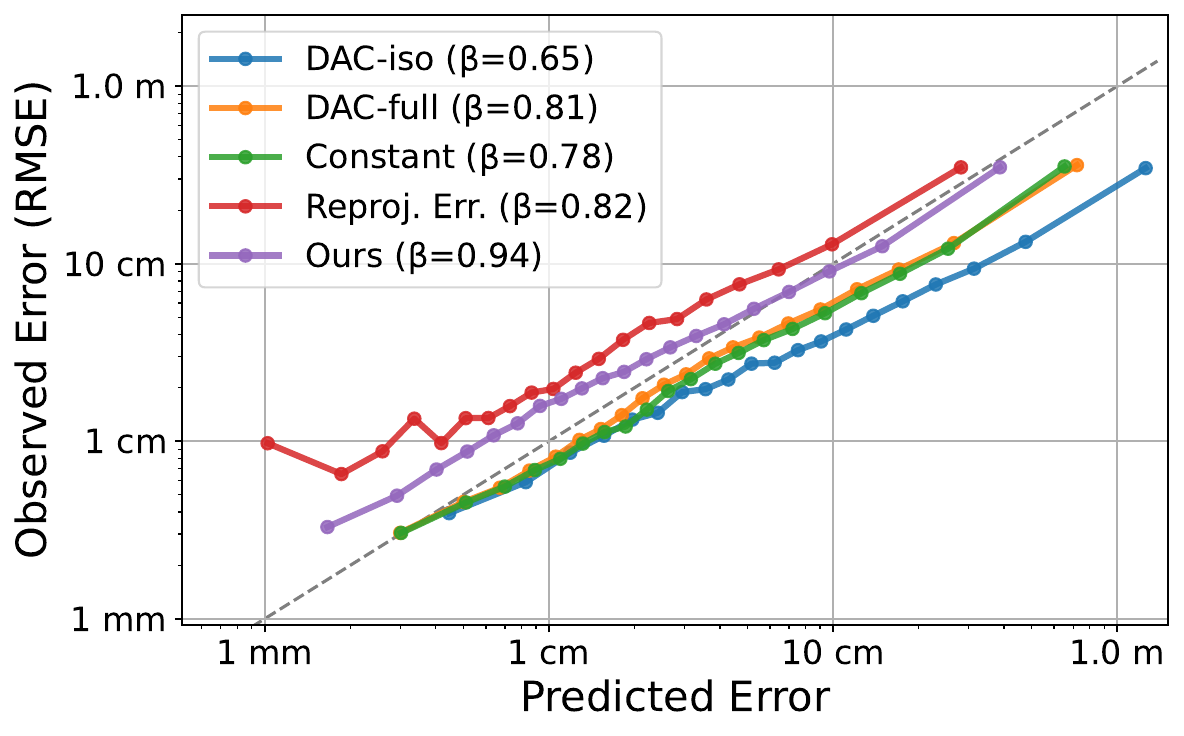}
    \caption{\textbf{Covariance Calibration.} We plot observed error against the ground truth vs.\ predicted uncertainty $\sqrt{\text{Tr}(\Sigma)}$. \ours\ is the closest to the ideal calibration $\beta{=}1$, shown as the $\color{gray}\text{dashed gray}$ line.}
    \label{fig:calibration_comparison}
\end{figure}

\paragraph{Results:} \cref{fig:calibration_comparison} shows our method ($\beta = 0.94$) which learns anisotropic 2D uncertainties of each keypoint achieves the closest agreement to the ideal unit slope, demonstrating that our covariances preserve a physical metric scale.

%% file: sec/7_conclusion.tex
\section{Conclusion}

We introduce RaCo, a lightweight neural network trained on perspective image crops only, addressing several challenges in keypoint detection.
Our method integrates a repeatable interest point detector with a differentiable ranker and a metric covariance estimator.
A ranker, trained to maximize repeatability, efficiently identifies the most valuable keypoints.
The covariance estimator provides a metric-scale measure of spatial uncertainty, valuable for downstream tasks.
Our approach achieves strong rotational robustness through a simple data augmentation strategy,
and experimental results validate our model's effectiveness in keypoint repeatability and two-view matching, particularly in scenes with large in-plane rotations. 
Ultimately, RaCo provides a simple yet effective strategy to detect robust interest points, rank keypoints, and quantify their metric covariance, making it a valuable building block for various computer vision systems.

%% file: sec/X_suppl.tex
\clearpage
\section*{Appendix}
\appendix

\section{Model Architecture}

\begin{figure*}[t]
  \centering
  \includegraphics[width=\linewidth]{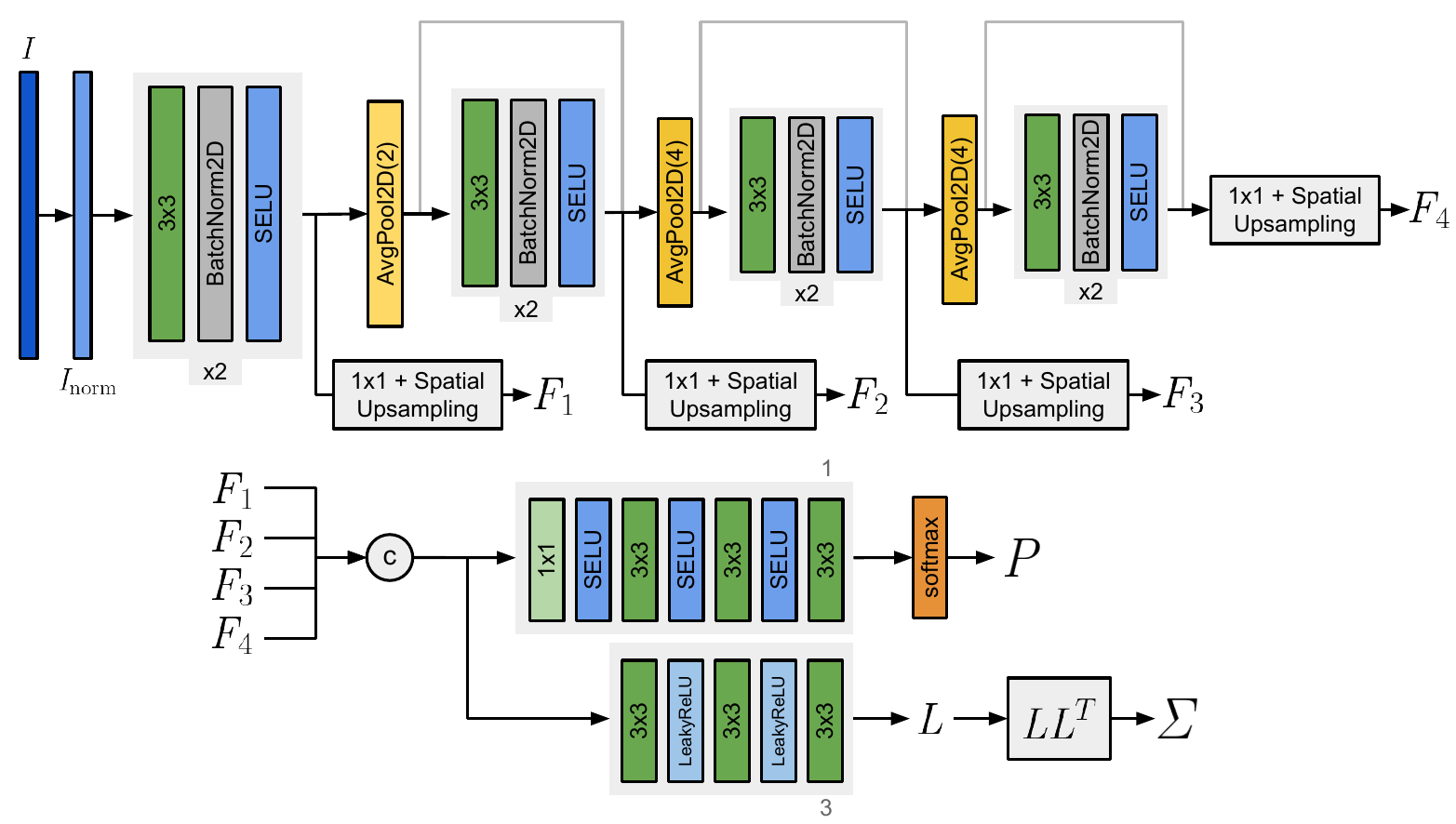}
  \caption{
  \textbf{Keypoint detector and covariance estimator model architecture.} 
  We use a multi-scale architecture with two heads, one which implements the detector and another which implements the covariance estimator.
  The network takes as input the RGB image $\textbf{\textit{I}}$, normalizes it to $I_{\text{norm}}$ and extracts multi-scale features $\textbf{\textit{F}}_i$, $i\in\{1,2,3,4\}$. 
  The detector head outputs the globally normalized heatmap $\textbf{\textit{P}}$. 
  The covariance estimator is a lightweight head which outputs the Cholesky decomposition map $\textbf{\textit{L}}$ over the whole image.
  }
  \label{fig:architecture_det_cov}
\end{figure*}

\begin{figure}[t]
    \centering
    \includegraphics[width=0.7\linewidth]{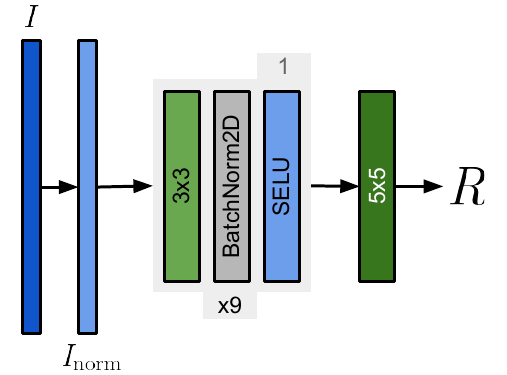}
    \caption{
    \textbf{Ranker model architecture.} The ranker is a simple standalone model that comprised of residual blocks. It takes as input the normalized image and outputs the single channel ranker score map $\textbf{\textit{R}}$.
    }
    \label{fig:architecture_rank}
\end{figure}

\cref{fig:architecture_det_cov} shows the architecture of \ours's keypoint detector and covariance estimator. 
We modify the architecture of ALIKED-N(16)~\cite{zhao2023aliked} by replacing the deformable convolution layers~\cite{xizhou2019deformable} with standard convolutions. 
We add the covariance estimator head which operates on the concatenated multi-scale features $\{\textbf{\textit{F}}_1, \textbf{\textit{F}}_2, \textbf{\textit{F}}_3, \textbf{\textit{F}}_4\}$ to produce the map of the Cholesky decomposition $\textbf{\textit{L}}$ from which $\mathbf{\mathit{\Sigma}}\in\mathbb{R}^{H\times W \times 2\times 2}$ is constructed. 
We observed that the covariance estimator head obtains a lower validation NLL loss when it shares the multi-scale features.
\cref{fig:architecture_rank} shows the architecture of our ranker module, which is simply a series of residual blocks that take as input the normalized RGB image and produce the ranker map $\textbf{\textit{R}}$.
The image is normalized using the standard ImageNet statistics, \ie, mean $\mu = [0.485, 0.456, 0.406]$ and standard deviation $\sigma = [0.229, 0.224, 0.225]$.

We train \ours\ in two stages, first by training the detector head and the multi-scale feature encoder with the training objective $\mathcal{L}_\text{detector}$. After this stage is complete, we freeze the detector head and multi-scale feature encoder's weights for the second stage.

In the second stage, we use the inference time settings for the detector, which means that we use the soft-argmax around a patch of the selected keypoint for subpixel sampling following \cite{edstedt2025dad, zhao2023aliked}. This is crucial for the training of the ranker and covariance estimator, as they require the inference time distribution of the keypoint correspondences and reprojection error. The ranker and covariance estimator can be trained in any order and combined later to obtain \ours, as they do not depend on each other.

\section{Evaluation Metrics}
\label{sec:eval_metrics_suppl}
Here we define the metrics reported in our evaluations in \cref{sec:experiments}.

\paragraph{Number of Matches:} We establish keypoint correspondences (matches) between two views in the following way: every keypoint in for example view $A$ is projected via the ground truth geometric transformation (either homography transformation or relative poses and ground truth depth) into the other view $B$. 
The same is done in the reverse direction and the nearest neighbors of every keypoint are computed. Matches/correspondences are declared to be pairs of keypoints which are mutual nearest neighbors, within a fixed matching radius.

\paragraph{Repeatability:} 
We define repeatability as the fraction of covisible points that have a corresponding detection in the other image, within an x-pixel reprojection error.
We compute the repeatability for each view separately and report the average repeatability of both views. We report this value at various matching thresholds.

\paragraph{Localization Error:} This is the average reprojection error of all matched keypoints in both views, reported in pixels. 
A lower localization error in our setting indicates a more spatially accurate keypoint detector.

\paragraph{Homography Corner Error AUC:}
For image pairs related by a ground truth homography, we evaluate the quality of the estimated homography by measuring the average corner reprojection error. 
Specifically, the corners of one image are warped using the estimated homography and compared against the ground truth corner locations. 
The mean Euclidean distance between the warped and true corner positions (in pixels) defines the homography corner error. 
Following prior work~\cite{superpoint}, we summarize performance by computing the Area Under the Curve (AUC) of the cumulative distribution of corner errors, up to different pixel thresholds. 
A higher AUC indicates more accurate and robust homography estimation.

\paragraph{Pose Error AUC:}
For image pairs with ground truth relative poses, we evaluate the accuracy of the estimated relative camera pose. 
The pose error is defined as the maximum of the angular error in rotation and the angular error in translation. 
We summarize performance by reporting the Area Under the Curve (AUC) of the cumulative distribution of pose errors, computed up to thresholds of $5^\circ$ and $10^\circ$. 
A higher AUC indicates more accurate and robust pose estimation.

\paragraph{Repeatability AUC:}
As described in \cref{ssec:rotationequivariance}, we evaluate rotation equivariance by measuring repeatability while rotating one of the two views in-plane over $360^\circ$. 
We then compute the Area Under the Curve (AUC) of the repeatability–rotation angle plot (see \cref{fig:rotation_eval}), where the rotation angles are normalized to the range $[0,1]$. 
An ideal rotation-equivariant detector achieves a repeatability AUC of $1$.

\paragraph{Multiview Triangulation Metrics:}
Following~\cite{schops2017multi}, we report \textbf{accuracy}, the fraction of reconstructed 3D points within a threshold of the ground truth surface (precision), and \textbf{completeness}, the fraction of ground truth points within a threshold of the reconstruction (recall).

\section{Supplementary Experimental Details}

\subsection{Homography Estimation}
\label{ssec:eval_protocol_hpatches}

Here we add more details about the evaluation in \cref{ssec:det_matching}. 
\cref{fig:repeatable_keypoint_comparison} contains some qualitative examples of repeatable keypoints for different detectors.

\paragraph{HPatches:}
We consider 540 image pairs from HPatches~\cite{balntas2017hpatches}. For comparable metrics, all images are resized so that the shorter side is 640 pixels.

\paragraph{DNIM:}
This dataset is based on the Archive of Many Outdoor Scenes (AMOS)~\cite{jacobs09webcamgis,jacobs07amos} and consists of sequences of images taken by one fixed camera per sequence, at various times across the day.
We randomly sample pairs of images from DNIM~\cite{zhou2016evaluating} such that there is a minimum time difference of half an hour between the capture time of the two images. 
The dataset contains images of the same scene taken at different times of the year, and we additionally create pairs by randomly sampling one image from each time of the year. 
This results in 428 random image pairs. To additionally evaluate the robustness of the keypoint detectors we augment the images with random homographies. 

\paragraph{Evaluation protocol:}
We first extract a fixed number of keypoints per view, 1024 keypoints for HPatches and 256 for DNIM. The images in DNIM are of low resolution, have low texture and the images at night suffer from shadow clipping. Forcing too many keypoints leads to spurious keypoint detections. We use the ground truth homography transformation to reproject points between views to compute the correspondences at a matching radius of 3px.
We estimate the homography using the Direct Linear Transformation (DLT)~\cite{hartley2003multiple} algorithm on all correspondences, as implemented in PoseLib~\cite{PoseLib}. For evaluation, we compute the homography corner error and report the area under the recall curve at thresholds of 1px and 3px.

\subsection{Relative Pose Estimation}

Here we add more details about the evaluation in \cref{ssec:det_matching}. 

\paragraph{MegaDepth1800:} This dataset is a subset of the test set of MegaDepth~\cite{li2018megadepth} introduced in \cite{lindenberger2023lightglue}. The dataset provides depth images, camera poses, and covisible image pairs from a large scale Structure-from-Motion (SfM) reconstruction of the scene. There are 4 diverse scenes in this subset. This data is used to project keypoints between views. We resize the images such that the longer side is 1600px long.

\paragraph{ETH3D-Two-View:} Based on the multi-view indoor \& outdoor ETH3D~\cite{schops2017multi} dataset, we create the ETH3D-Two-View subset. It contains covisible image pairs, ground truth depth images from a LiDAR scanner and ground truth camera poses. Our subset consists of 1171 image pairs across 13 scenes. We resize the images such that the longer side is 1024px long.

\paragraph{Evaluation protocol:} We first extract a fixed number of keypoints per view, 2048 keypoints for both datasets. We use the ground truth camera poses and depth to reproject points between views and compute the correspondences at a matching radius of 5px.
We employ a robust pose estimation pipeline using RANSAC~\cite{fischler1981random} from the PoseLib library~\cite{PoseLib}. We individually optimize the inlier threshold for each method to ensure a fair evaluation of performance. We select the optimal threshold from the set of $\{0.5,1.0,1.5,2.0,2.5,3.0\}$. The pose error is measured as the maximum angular difference between the ground-truth and estimated rotation and translation. We report the area under the recall curve (AUC) at angular thresholds of $5^\circ$ and $10^\circ$.

\subsection{Rotation Equivariance}
We provide \textbf{video examples} in the supplementary material demonstrating our model's rotation equivariance evaluated in \cref{ssec:rotationequivariance}. Across a full $360^\circ$ rotation of the second view, our model consistently produces more matches and exhibits greater matching stability compared to baselines.

\begin{figure}[t] 
    \centering
    \includegraphics[width=\linewidth]{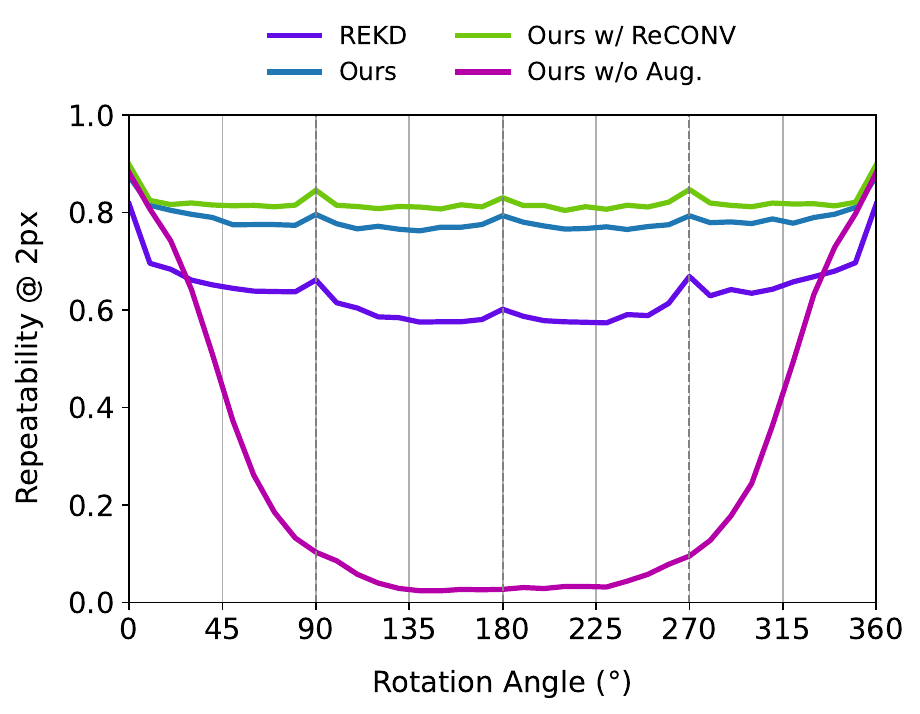}
    \caption{\textbf{Addtional rotation ablations on HPatches~\cite{balntas2017hpatches}.} We plot the repeatability@2px over the rotation angle between image pairs just as in \cref{fig:rotation_eval}. We ablate our method against our method with equivariant convolutions~\cite{weiler2019general}, ours without rotation agumentations, and REKD~\cite{lee2022self}, a recent baseline which uses rotationally equivariant convolutions (ReCONVs). Adding equivariant convolutions only adds minor stability at large computational cost. The gap between our model and REKD is large, even without equivariant convolutions, suggesting that the proposed rotation augmentations are effective.
    }
    \label{fig:rotation_repeatability_ablation}
\end{figure}

In \cref{fig:rotation_repeatability_ablation} we show the results of the same evaluation as in \cref{ssec:rotationequivariance} for the design choice of not using special architectures such as rotationally equivariant convolutions~\cite{cesa2022program}. We include REKD~\cite{lee2022self} in our comparison, it is a model that uses rotationally equivariant convolutions based on \cite{weiler2019general}.

Looking at \cref{fig:rotation_eval} and \cref{fig:rotation_repeatability_ablation}, detectors often exhibit some periodicity at a frequency of $90^\circ$, including ALIKED~\cite{zhao2023aliked} and REKD~\cite{lee2022self}. This can be attributed to two factors: a) as opposed to our strong rotation augmentations, learned detectors are often trained with rotation augmentations at $90^\circ$ intervals and they are more equivariant specifically at these rotation angles, and b) the interpolation artifacts from rotating a grid of square pixels disappear at rotation angles that are multiples of $90^\circ$. This has been studied in Fig. 7 of \cite{lee2022self}.

\subsection{Keypoint Ranking}
In \cref{ssec:rankingeval}, we evaluate keypoint ordering by extracting a fixed number of keypoints per view: 1024 for HPatches~\cite{balntas2017hpatches} and 2048 for MegaDepth1800~\cite{li2018megadepth, sun2021loftr}. To assess performance at a given keypoint budget $n$, we sort the extracted points in descending order by either detector or ranker score and consider only the top $n$ points for matching.

\subsection{Multiview Triangulation}
\begin{figure}[t]
    \centering
    \includegraphics[width=\linewidth]{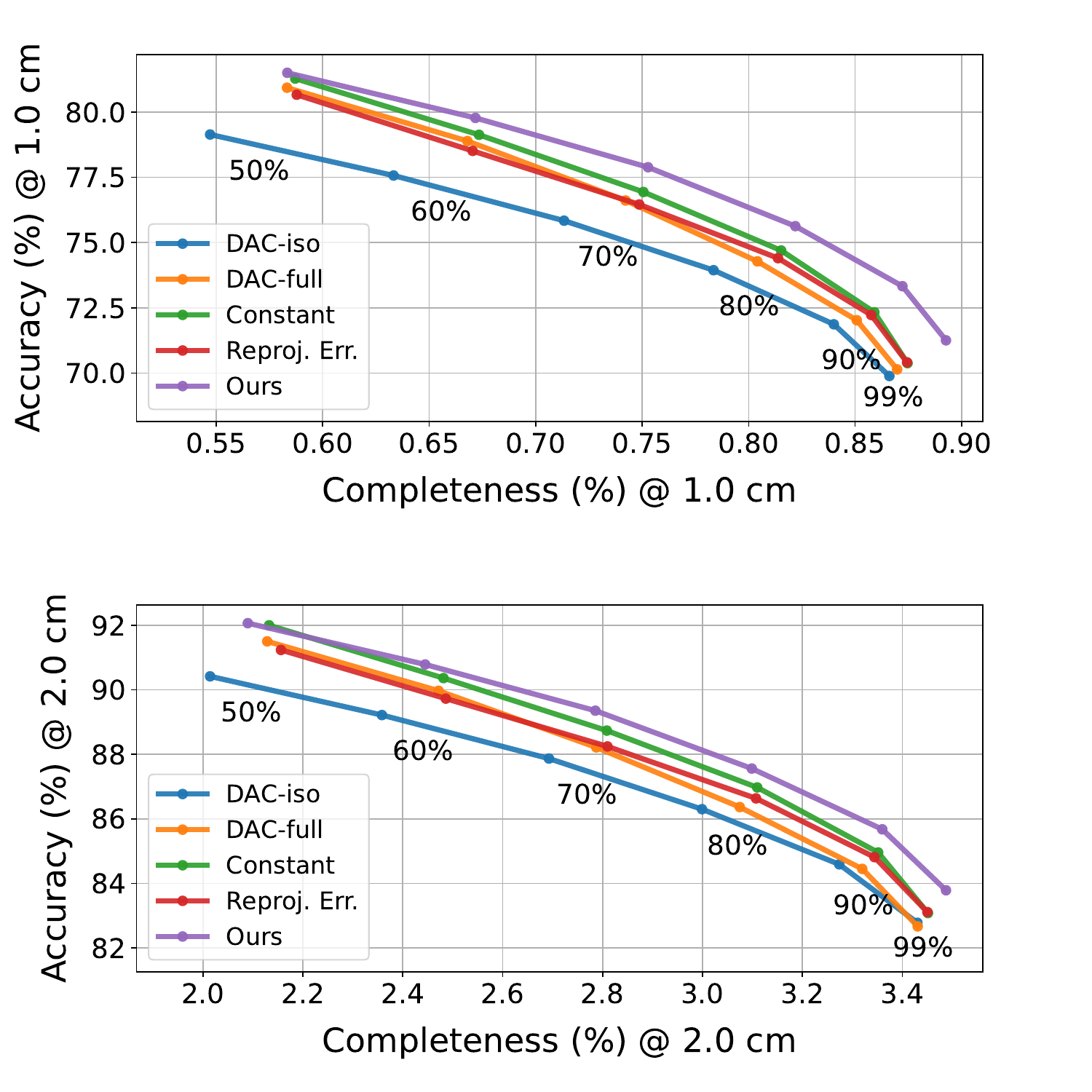}
    
    \caption{
    \textbf{3D triangulation on ETH3D~\cite{schops2017multi}.} We plot the triangulated point clouds' accuracy against completeness within 1 cm and 2 cm. Each point corresponds to the fraction of the original point cloud retained after filtering based on different spatial uncertainty metrics.
    }
    \label{fig:filtering_acc_vs_completness_more}
\end{figure}

\cref{fig:filtering_acc_vs_completness_more} provides the results of \cref{ssec:coveval} at two more coarse thresholds of 1 cm and 2 cm.

\subsection{Multiview Triangulation Detector Evaluation}
\paragraph{Setup:} We evaluate the quality of the keypoints on a downstream task of multiview triangulation. We follow the setup of \cref{ssec:coveval} and extract keypoints and correspondences on the ETH3D~\cite{schops2017multi} dataset.
We triangulate the matches to form a pointcloud on which we run 3D point only bundle adjustment, where the camera parameters are held constant.
For keypoint extraction, we resize the images such that the largest side is 1024px long just as in~\cref{ssec:det_matching}, and extract 4096 keypoints per image.

\paragraph{Baselines:} We compare ALIKED~\cite{zhao2023aliked}, DaD~\cite{edstedt2025dad} and SuperPoint~\cite{superpoint} against our model with their default settings. We compare the F1 score computed using the accuracy and completeness following \cite{schops2017multi} and \cref{sec:eval_metrics_suppl} of the point cloud at thresholds of 0.5 cm, 1 cm, and 2 cm. We run our method with the learned covariances and use it to weight the reprojection errors in bundle adjustment.

\paragraph{Results:} \cref{tab:f1_score_results} shows that our model is competitive with the other learned methods on accuracy and achieves the highest completeness over those thresholds.

\begin{table}[t]
\centering
\begin{tabular}{@{}lccc@{}}
\toprule
 & \multicolumn{3}{c}{F1 Score (\%) @ $\tau$ cm} \\
\cmidrule(l r){2-4} 
& $\tau = 0.5$ & $\tau = 1.0$ & $\tau = 2.0$ \\
\midrule
SuperPoint~\cite{superpoint} & 0.30 & 1.66 & 6.50 \\
ALIKED~\cite{zhao2023aliked} & 0.29 & 1.58 & 6.01 \\
DaD~\cite{edstedt2025dad} & \cellcolor{tabsecond}0.32 & \cellcolor{tabsecond}1.71 & \cellcolor{tabsecond}6.69 \\
Ours & \cellcolor{tabfirst}0.33 & \cellcolor{tabfirst}1.76 & \cellcolor{tabfirst}6.71 \\
\bottomrule
\end{tabular}
\caption{
F1 scores (\%) at different thresholds for various methods.
The \colorbox{tabfirst}{best} and \colorbox{tabsecond}{second best} results for each metric are colored.
}
\label{tab:f1_score_results}
\end{table}

\section{Qualitative Examples}
\label{sec:qualitative_suppl}

In \cref{fig:hpatches_qualitative} and \cref{fig:megadepth_eth3d_qualitative} we provide some qualitative examples of {\ours}'s outputs. We visualize the detector scoremap, the ranker scoremap and an interpretable map of the uncertainty estimate on images from datasets used in \cref{ssec:det_matching}.

We construct the rightmost maps in \cref{fig:hpatches_qualitative} and \cref{fig:megadepth_eth3d_qualitative} from the covariance map, $\mathbf{\mathit{\Sigma}} \in \mathbb{R}^{H\times W\times 2\times 2}$, for each pixel. 
Each pixel's color is determined by the angle of the major axis of its covariance ellipse.
The color intensity is weighted by $|\mathbf{\Sigma}|$, a measure of the pixel's total spatial uncertainty, such that regions of higher uncertainty appear whiter.

\cref{fig:repeatable_keypoint_comparison} contains some qualitative examples of repeatable keypoints for different detectors.

Qualitative examples of our ranker module are demonstrated on HPatches~\cite{balntas2017hpatches} in \cref{fig:ranker_qualitative}. By independently reordering keypoints in each view based on their ranking scores, the ranker significantly boosts repeatability at the illustrated budgets of 128 and 256 keypoints.

\input{figures/detector_qualitative/v1/all_examples}

\input{figures/repeatable_qualitative/v1/all_repeatable}

\clearpage
\input{figures/ranker_qualitative/all_ranker_qualitative}

%% file: figures/detector_qualitative/v1/all_examples.tex
\newlength{\detectorimgwidth}
\setlength{\detectorimgwidth}{0.199\textwidth}

\newcolumntype{B}[1]{>{\centering\arraybackslash}b{#1}}

\newcommand{\imageexample}[1]{%
    \begin{tabular}{cccc}
        \includegraphics[width=\detectorimgwidth]{figures/detector_qualitative/v1/#1_cropped_img_keypoints.jpg} &
        \includegraphics[width=\detectorimgwidth]{figures/detector_qualitative/v1/#1_keypoint_log_heatmap.jpg} &
        \includegraphics[width=\detectorimgwidth]{figures/detector_qualitative/v1/#1_ranker_heatmap.jpg} &
        \includegraphics[width=\detectorimgwidth]{figures/detector_qualitative/v1/#1_eigenvector_angle_weighted.jpg} \\
    \end{tabular}
}

\newcommand{\imageexampleAndSpace}[1]{%
    \imageexample{#1} \\[0.5em]
}

\newcommand{\addlegends}{%
    \begin{tabular}{cccc}
        \raisebox{0pt}[0pt][0pt]{\includegraphics[width=\detectorimgwidth, height=0pt]{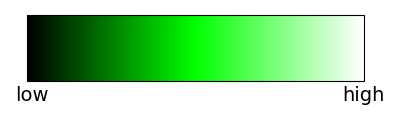}}&
        \includegraphics[width=\detectorimgwidth]{figures/detector_qualitative/v1/legend/keypoint_legend.jpg} &
        \includegraphics[width=\detectorimgwidth]{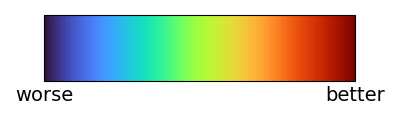} &
        \includegraphics[width=\detectorimgwidth]{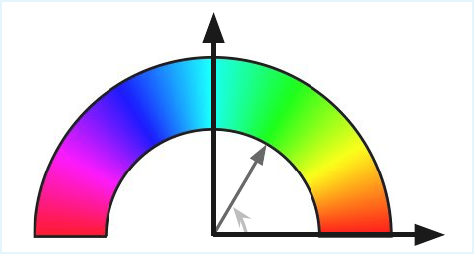} \\
    \end{tabular} \\[0.5em]
}

\newcommand{\createfigure}[5]{
    \clearpage
    \vfill
    \begin{figure*}[p] 
        \centering
        \setlength{\tabcolsep}{0pt}
        
        \begin{tabular}{B{\detectorimgwidth} B{\detectorimgwidth} B{\detectorimgwidth} B{\detectorimgwidth}}
            Keypoints &
            Detector Scoremap &
            Ranker Scoremap &
            \shortstack{Covariance Ellipse \\ Major Axis Angle} \\
        \end{tabular}
        
        \addlegends
        \begin{tabular}{c} 
            #1
        \end{tabular}
        \caption{\textbf{\ours's outputs on #2.} #3}
        \label{#4}
    \end{figure*}
}

\createfigure{
    \imageexampleAndSpace{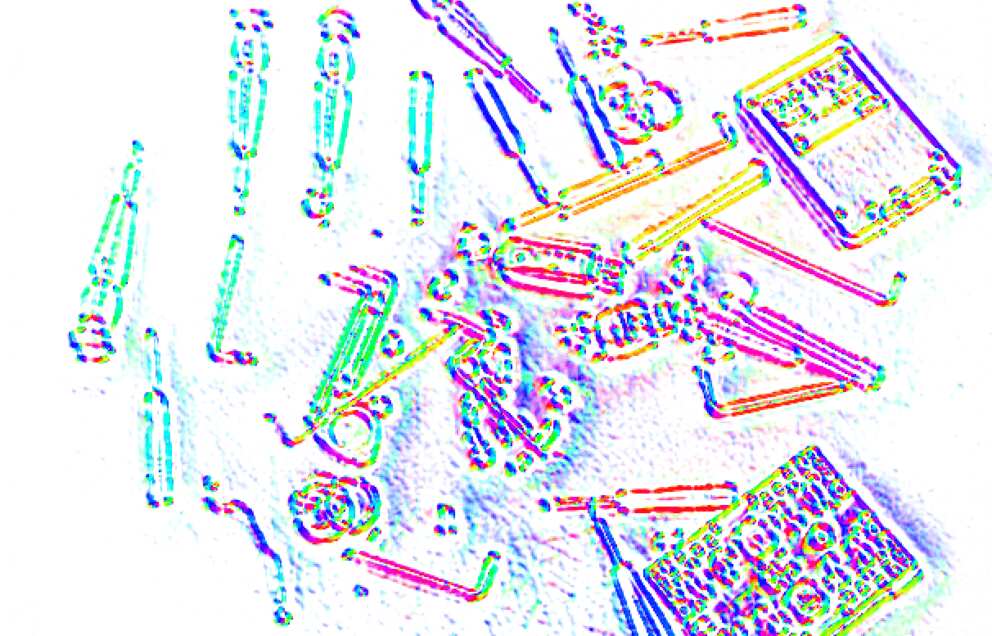}
    \imageexampleAndSpace{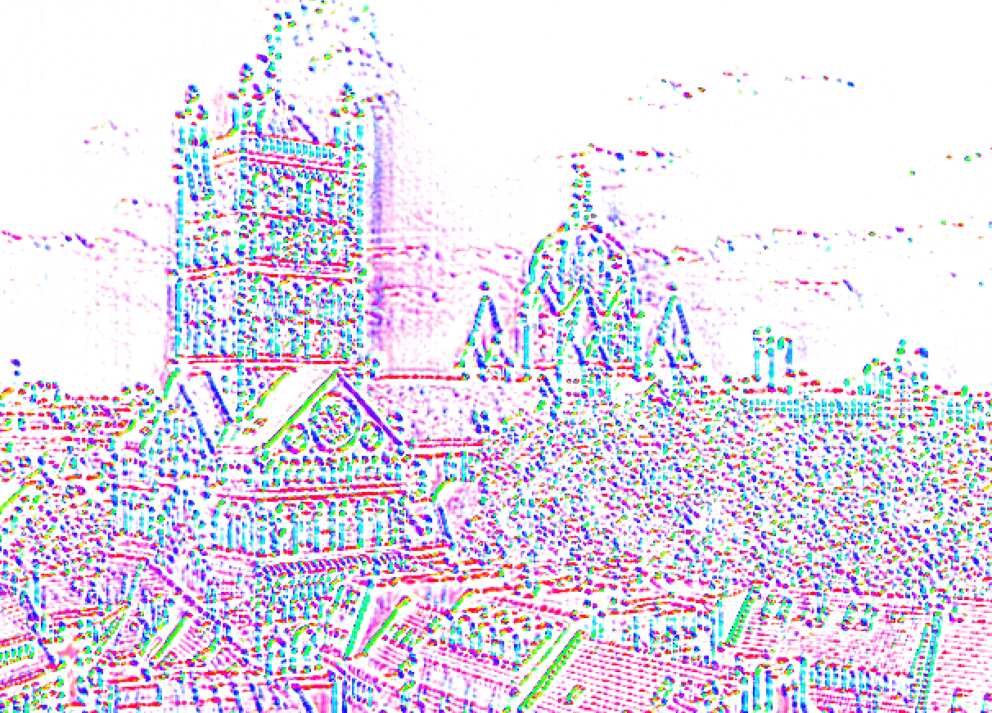}
    \imageexampleAndSpace{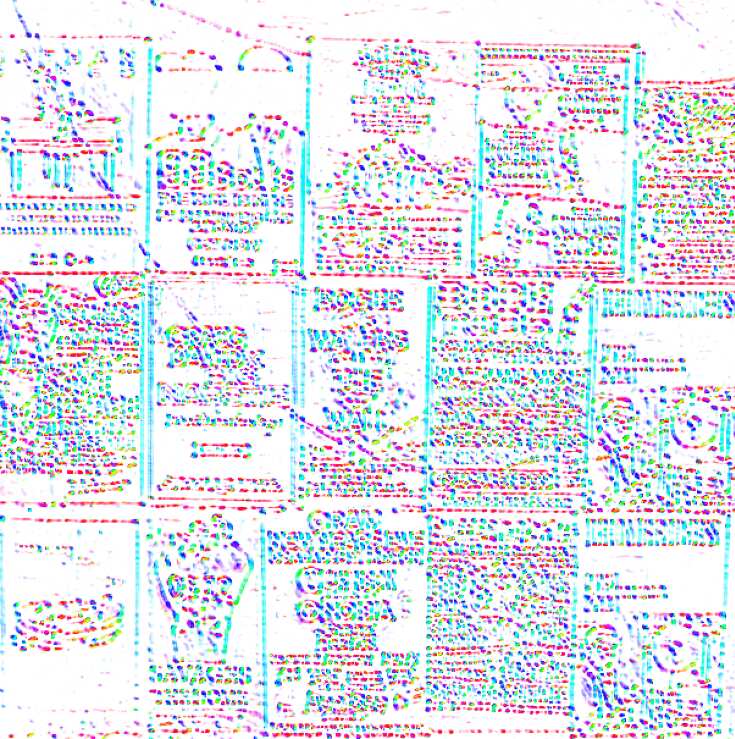}
    \imageexampleAndSpace{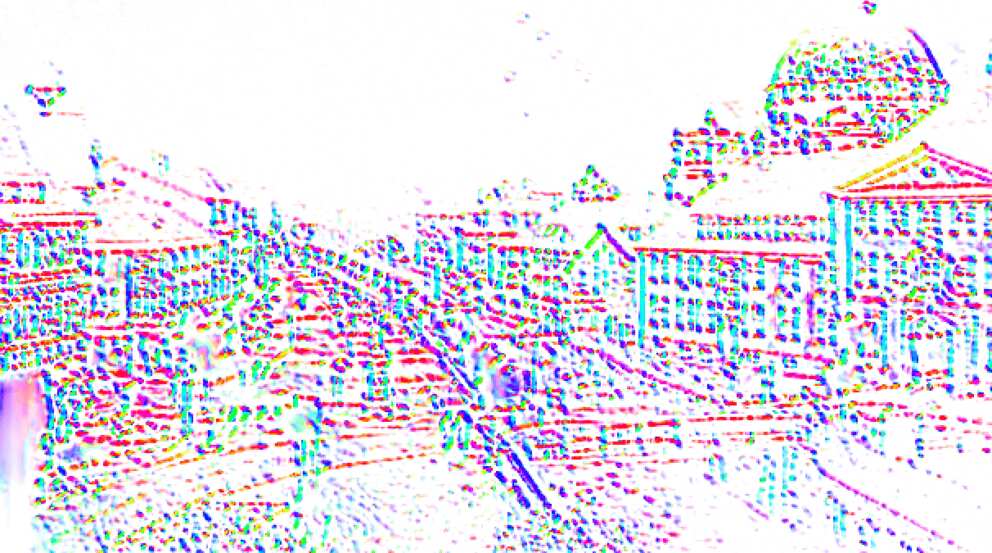}
    \imageexampleAndSpace{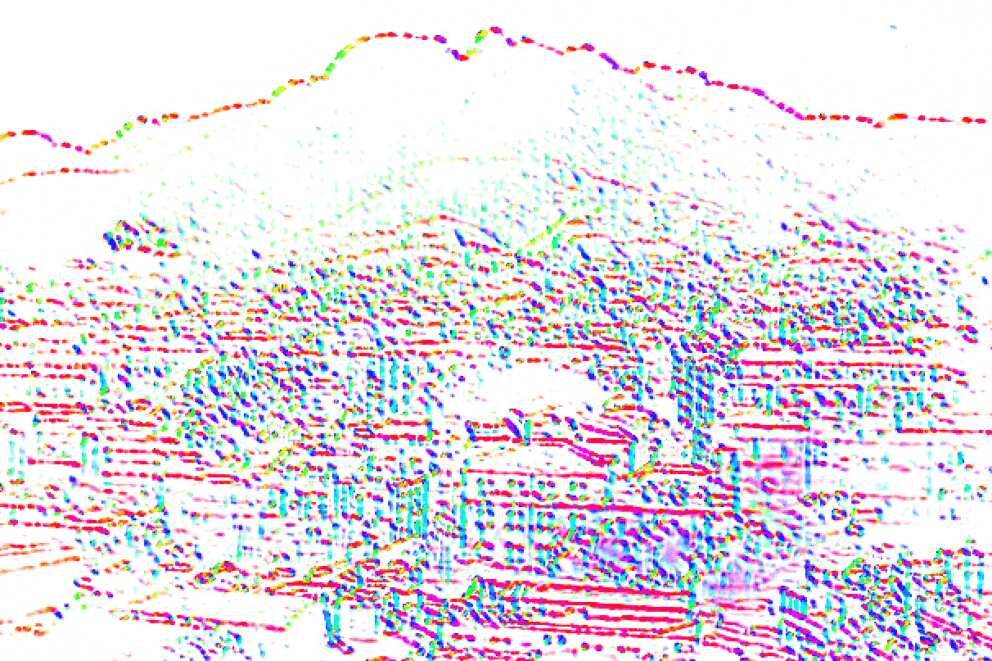}
    \imageexample{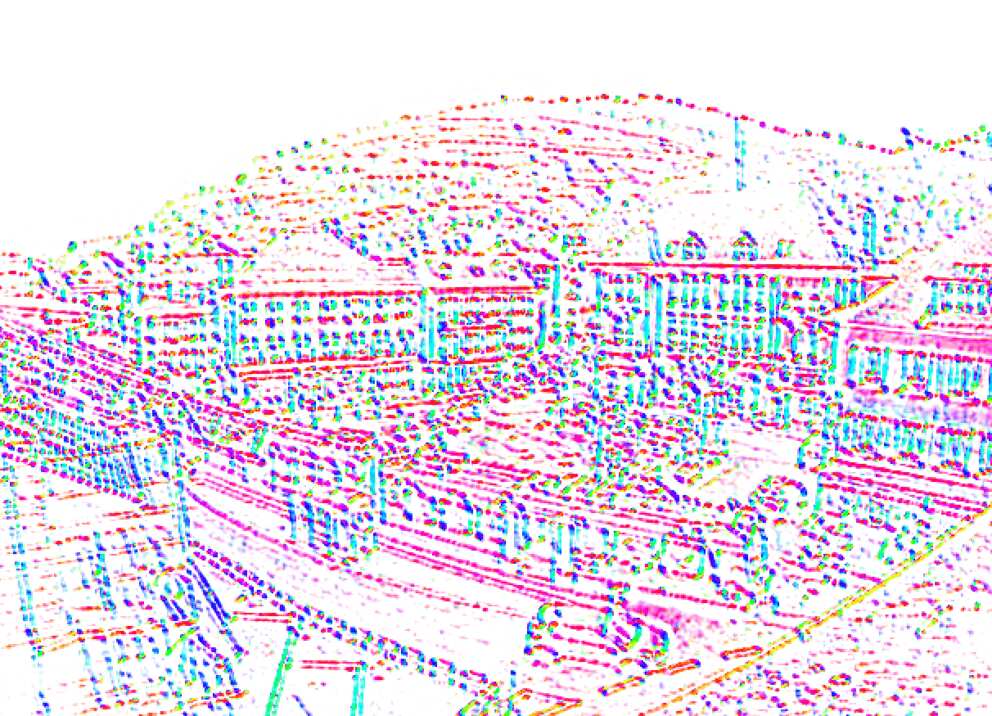}
}{HPatches and DNIM}{
    (left) Keypoints overlayed on the input image (middle left) The detector probability score map (middle right) The ranker score map (right) The weighted map of the angle of the major axis of the estimated covariance ellipsoid described in \cref{sec:qualitative_suppl}.
    Our detector learns corner like features that maximize the repeatability objective.
    The ranker learns an ordering of these keypoints, notice how more prominent corners are assigned a higher ranker score, less prominent corners are ranked lower, followed by edge like features. Textureless regions are assigned the lowest rank as they are the least likely to be matched.
}{fig:hpatches_qualitative}{
}

\createfigure{
    \imageexampleAndSpace{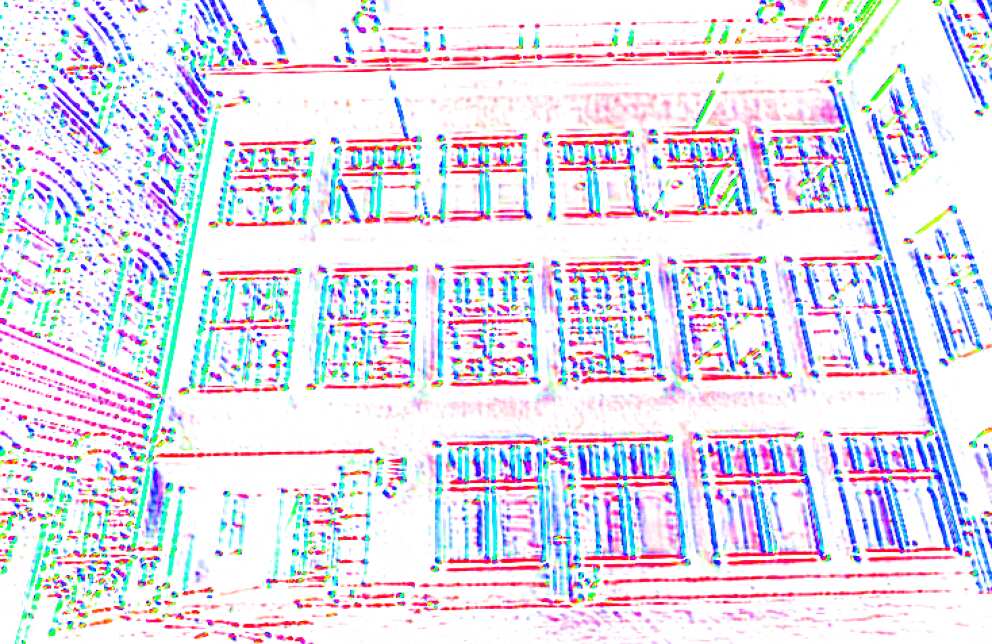}
    \imageexampleAndSpace{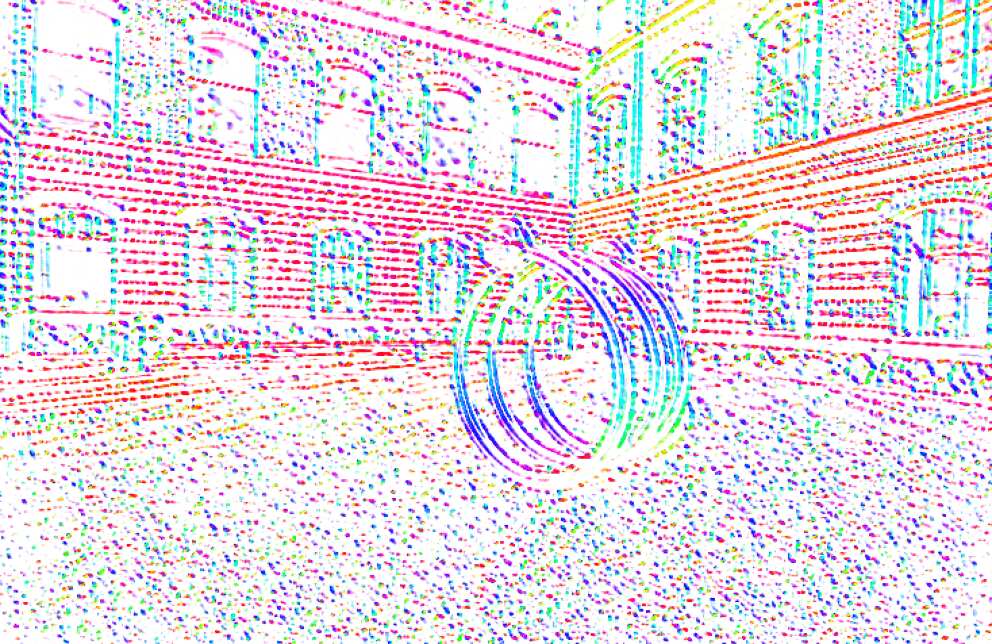}
    \imageexampleAndSpace{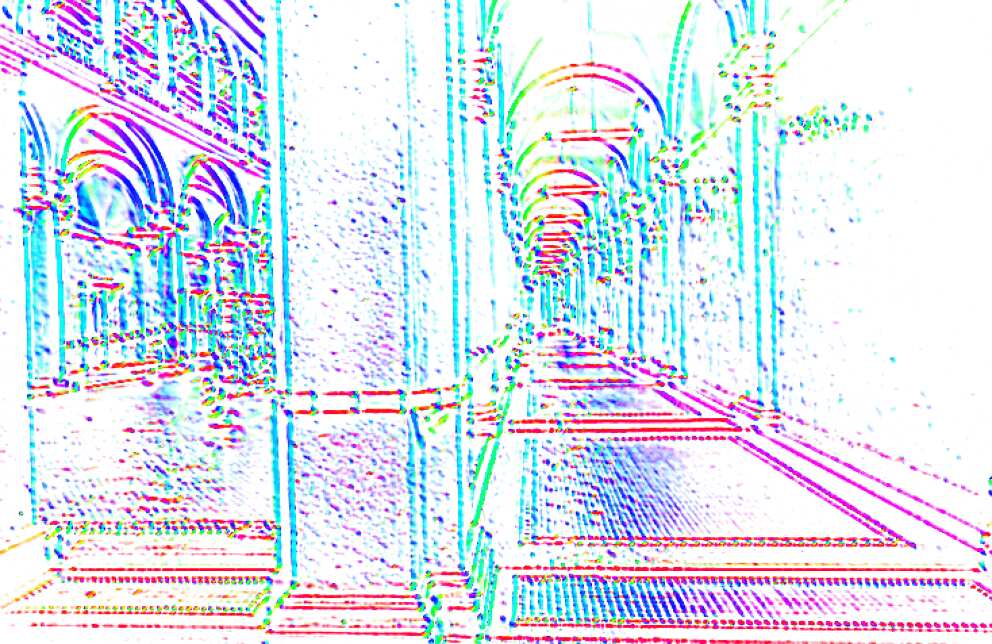}
    \imageexampleAndSpace{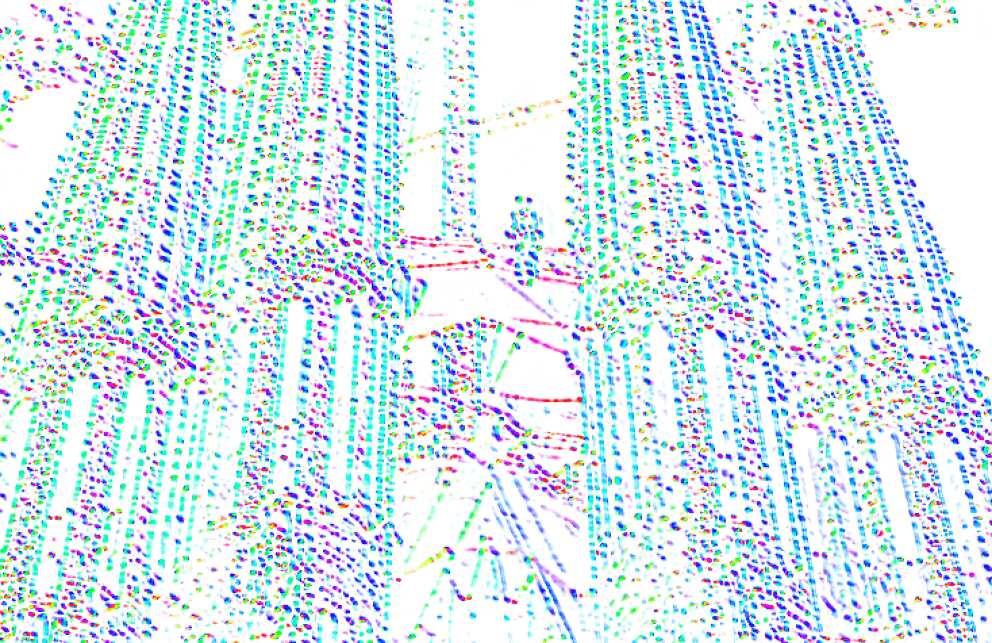}
    \imageexampleAndSpace{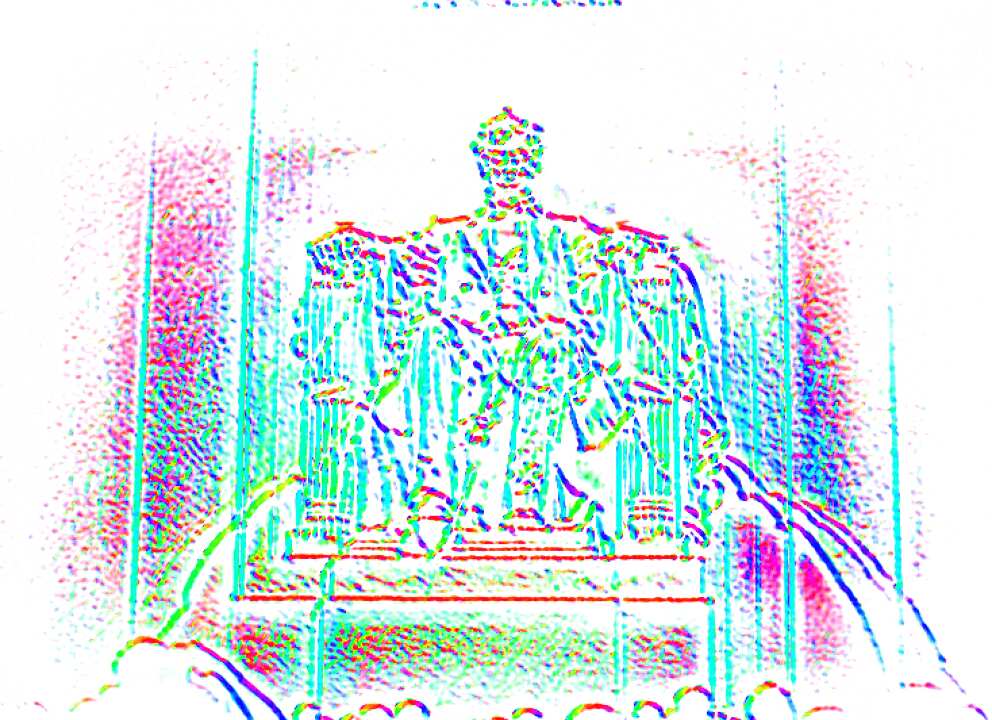}
    \imageexampleAndSpace{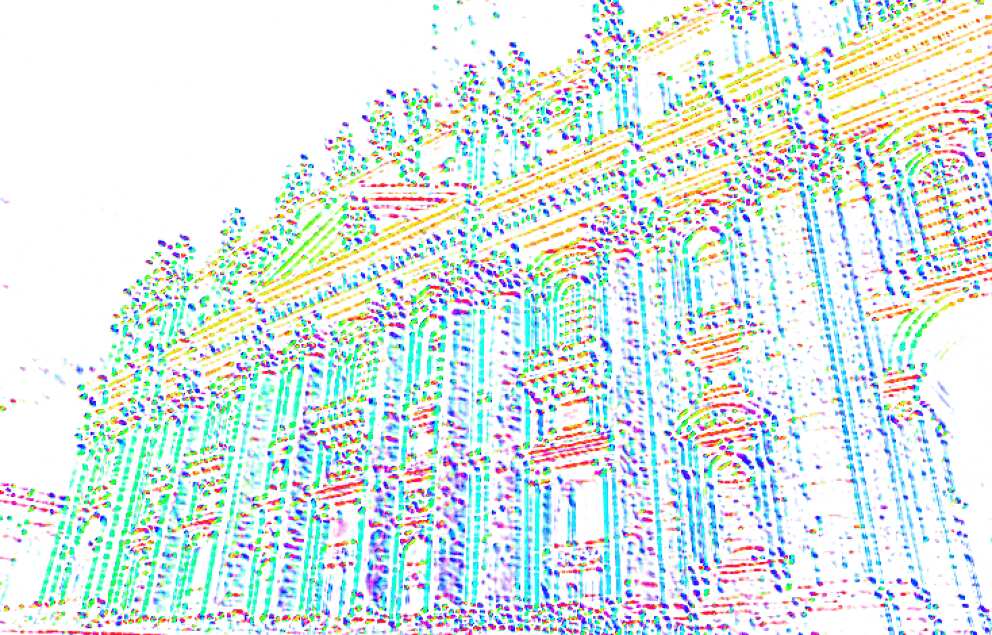}
    \imageexample{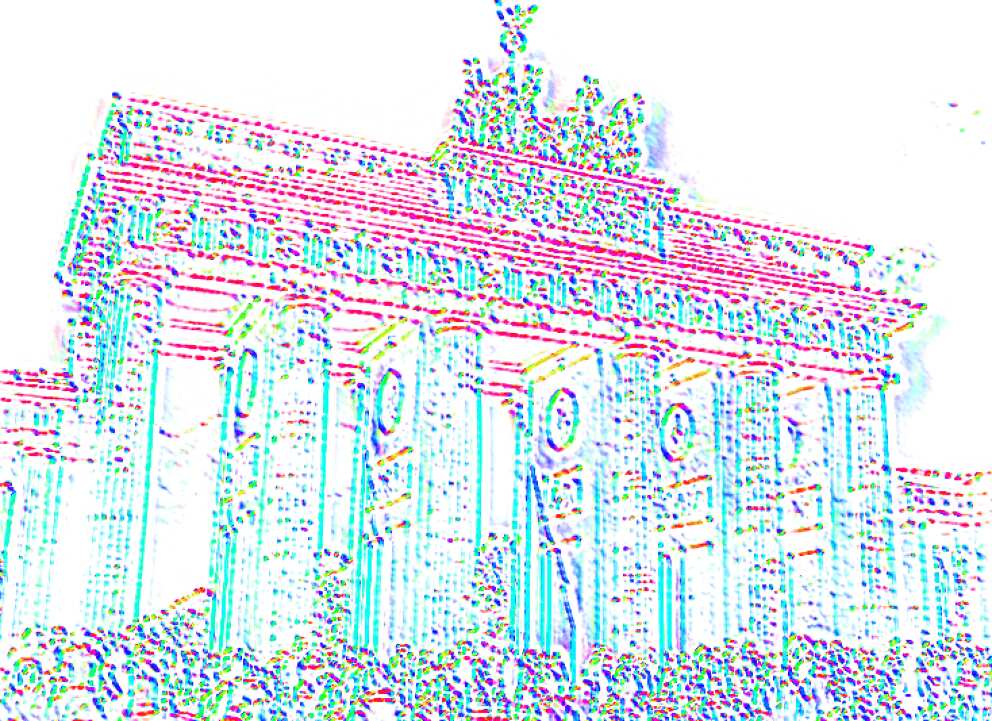}
}{MegaDepth and ETH3D}{
    (left) Keypoints over the image (middle left) Detector score map (middle right) Ranker score map (right) The weighted map of the angle of the major axis of the estimated covariance ellipsoid described in \cref{sec:qualitative_suppl}.
    The estimated covariances are very interpretable, in regions of low texture the rightmost plot is more white as the uncertainty in those regions is higher.
    On edge, corner and blob like image features, the uncertainty is much lower.
    Further, the angle of the major axis of the estimated ellipsoid follows the angle of the edge.
    This is clearly seen in the second row around the metallic helical like structure.
}{fig:megadepth_eth3d_qualitative}{
}

%% file: figures/repeatable_qualitative/v1/all_repeatable.tex
\newlength{\repeatableimgwidth}
\setlength{\repeatableimgwidth}{0.17\textwidth}

\newcommand{\repeatableimgpath}{figures/repeatable_qualitative/v1/}

\newcommand{\addimagegrid}[1]{%
    \begin{tabular}{B{\repeatableimgwidth}B{\repeatableimgwidth}B{\repeatableimgwidth}B{\repeatableimgwidth}}
        \includegraphics[width=\repeatableimgwidth]{\repeatableimgpath #1/#1_superpoint_view0.jpg} &
        \includegraphics[width=\repeatableimgwidth]{\repeatableimgpath #1/#1_aliked_view0.jpg} &
        \includegraphics[width=\repeatableimgwidth]{\repeatableimgpath #1/#1_dad_view0.jpg} &
        \includegraphics[width=\repeatableimgwidth]{\repeatableimgpath #1/#1_ours_view0.jpg} \\
        \includegraphics[width=\repeatableimgwidth]{\repeatableimgpath #1/#1_superpoint_view1.jpg} &
        \includegraphics[width=\repeatableimgwidth]{\repeatableimgpath #1/#1_aliked_view1.jpg} &
        \includegraphics[width=\repeatableimgwidth]{\repeatableimgpath #1/#1_dad_view1.jpg} &
        \includegraphics[width=\repeatableimgwidth]{\repeatableimgpath #1/#1_ours_view1.jpg} \\
    \end{tabular}
}

\begin{figure*}[p]
    \centering
    \setlength{\tabcolsep}{0pt} 
    \renewcommand{\arraystretch}{0} 

        \addimagegrid{example0} \\[0.5em]
        \addimagegrid{example1} \\[0.5em]
        \addimagegrid{example2} \\[0.5em]
        \addimagegrid{example3}

    \caption{
    \textbf{Qualitative examples of two-view matching.}
    The repeatable points ($\color{lime}\bullet$), unmatched points ($\color{red}\bullet$) and the non-covisible points ($\color{blue}\bullet$) are showed for examples from HPatches~\cite{balntas2017hpatches} and MegaDepth~\cite{li2018megadepth} from the evaluation in \cref{ssec:det_matching}. We additionally report the repeatability of each detector. Despite being trained solely on image pairs of perspective image crops, our model generalizes to images with real viewpoint and illumination changes.
    }
    \label{fig:repeatable_keypoint_comparison}
\end{figure*}

%% file: figures/ranker_qualitative/all_ranker_qualitative.tex
\newlength{\rankerimagewidth}
\setlength{\rankerimagewidth}{0.8\textwidth} 

\clearpage
\vfill
\begin{figure*}[p]
    \centering
    \includegraphics[width=\rankerimagewidth]{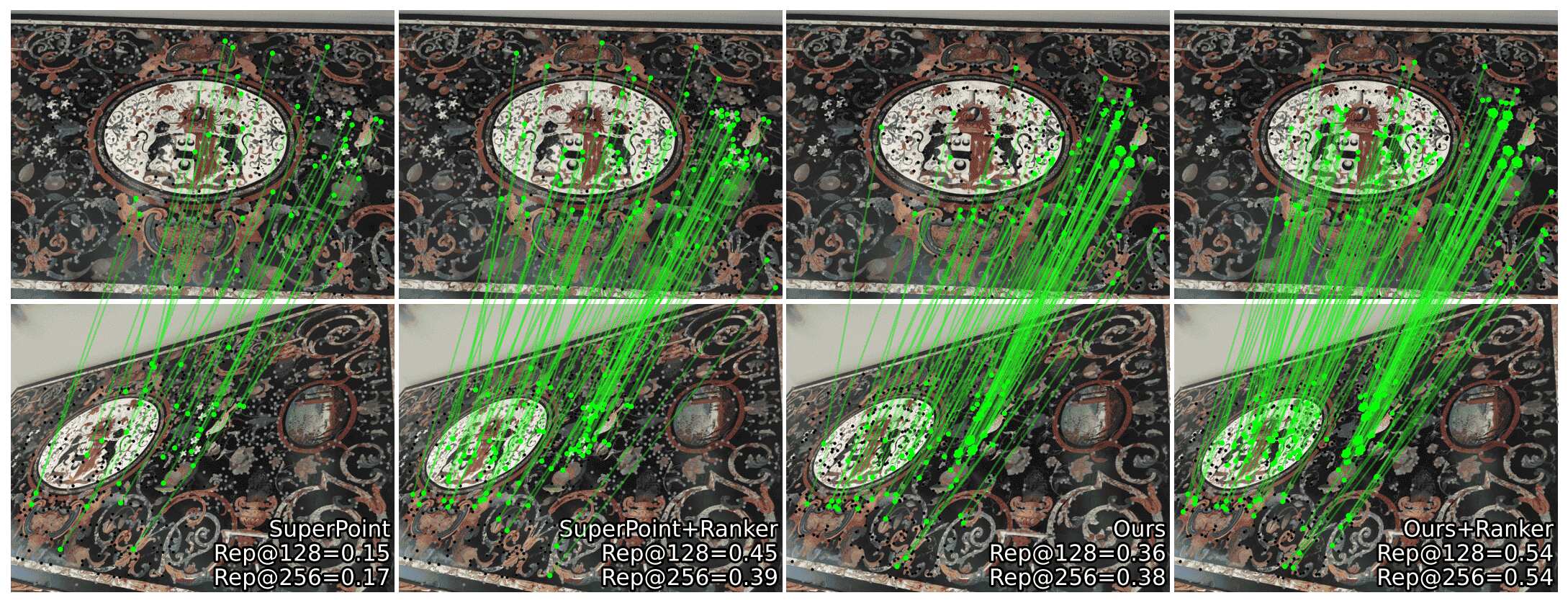} \\
    \includegraphics[width=\rankerimagewidth]{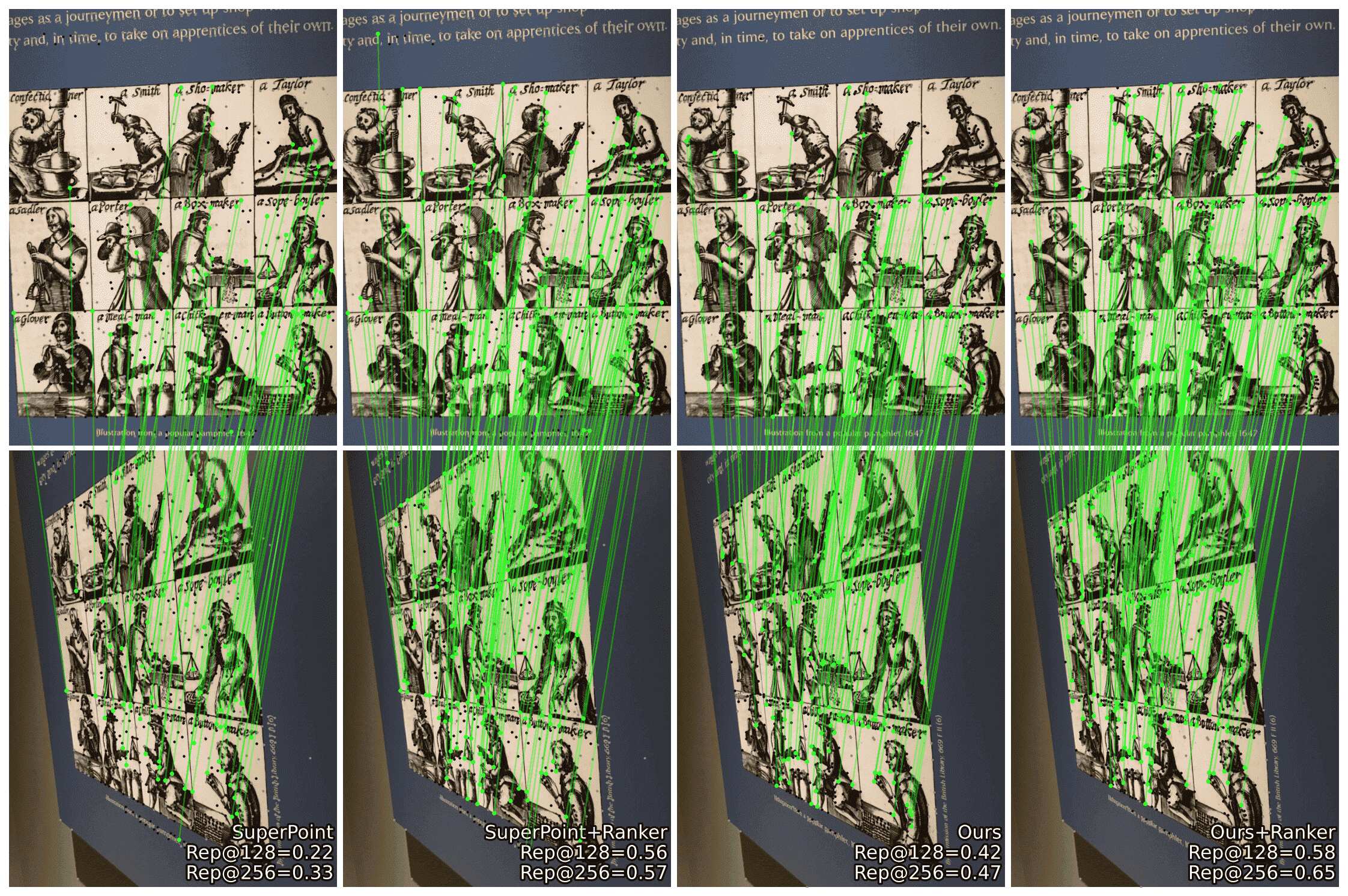} \\
    \includegraphics[width=\rankerimagewidth]{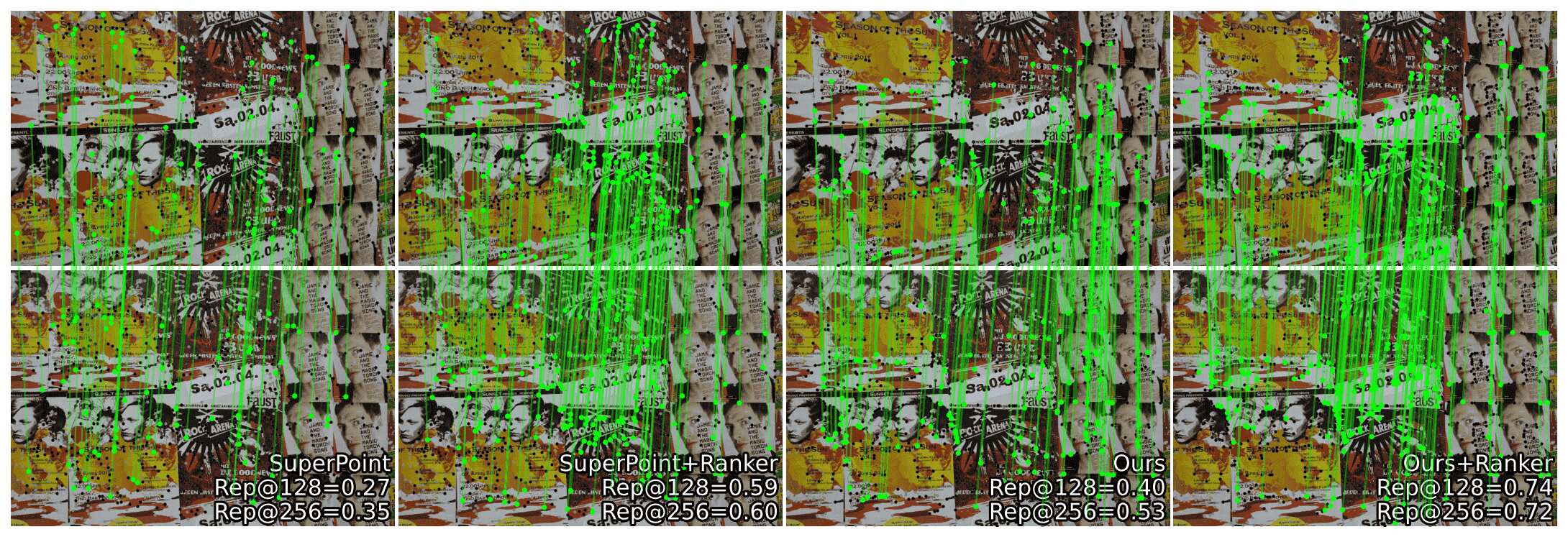} \\
    \caption{
    \textbf{Qualitative Analysis of Keypoint Ranking.}
    We provide a qualitative evaluation of our keypoint ranking method by visualizing keypoints and their matches for three examples from HPatches~\cite{balntas2017hpatches} from the evaluation in \cref{ssec:rankingeval}. 
    With a fixed budget of 256 keypoints per view, ordering by our ranker scores ($\mathbf{r}$) significantly increases the number of matches compared to ordering by detector scores ($\mathbf{p}$). 
    This demonstrates the ranker's ability to prioritize \textbf{matchable} keypoints ($\color{lime}\bullet$ at a budget of 256) while de-prioritizing \textbf{unmatchable} ones ($\color{black}\bullet$). The keypoints that would be matched had the full budget been considered are also shown ($\color{gray}\bullet$).
    We also report repeatability scores at keypoint budgets of 128 and 256.
    }
    \label{fig:ranker_qualitative}
\end{figure*}